\documentclass[lettersize,journal]{IEEEtran}
\usepackage{amsmath}
\usepackage{algorithm}
\usepackage{algorithmic}
\usepackage{array}
\usepackage[caption=false,font=normalsize,labelfont=sf,textfont=sf]{subfig}
\usepackage{textcomp}
\usepackage{stfloats}
\usepackage{url}
\usepackage{verbatim}
\usepackage{graphicx}
\usepackage{cite}
\usepackage{svg}
\usepackage{booktabs}
\usepackage{multirow}
\usepackage{pifont} 
\newcommand{\cmark}{\ding{51}} % check mark
\newcommand{\xmark}{\ding{55}} % cross mark
\usepackage{makecell}
\usepackage{tabularx}    % 导言区加载
\usepackage{tabulary}
\usepackage{xcolor}
\usepackage{enumitem}
\begin{document}

%\title{SandWorm: An Event-based Vibrational Visuotactile Sensor enabled Screw-Peristaltic Driven Robot for Granular Media Exploration}
\title{SandWorm: Event-based Visuotactile Perception with Active Vibration for Screw-Actuated Robot in Granular Media}

\author{ Shoujie Li$^*$, Changqing Guo$^*$,  Junhao Gong, Chenxin Liang, Wenhua Ding, Wenbo Ding

\thanks{*These authors contributed equally to this
work.}
\thanks{This work was supported by the National Key R\&D Program of China grant (2024YFB3816000), Guangdong Innovative and Entrepreneurial Research Team Program (2021ZT09L197), Tsinghua Shenzhen International Graduate School-Shenzhen Pengrui Young Faculty Program of Shenzhen Pengrui Foundation (No. SZPR2023005) and Meituan.  
(Corresponding author: Wenbo Ding, ding.wenbo@sz.tsinghua.edu.cn) }   %
\thanks{Shoujie Li, Changqing Guo,  Junhao Gong, Chenxin Liang, Wenhua Ding, Wenbo Ding are with Tsinghua Shenzhen International Graduate School,  Shenzhen 518055, China. (email: \{lsj20, gcq24, gongjh24, liangcx23, dingwh24\}@mails.tsinghua.edu.cn, ding.wenbo@sz.tsinghua.edu.cn))}
\thanks{Shoujie Li is also with the School of Mechanical and Aerospace Engineering, Nanyang Technological University, Singapore 639956, Singapore.}}

\markboth{Journal of \LaTeX\ Class Files,~Vol.~14, No.~8, August~2021}%
{Shell \MakeLowercase{\textit{et al.}}: A Sample Article Using IEEEtran.cls for IEEE Journals}


\maketitle

\begin{abstract}
Perception in granular media remains challenging due to unpredictable particle dynamics. To address this challenge, we present SandWorm, a biomimetic screw‑actuated robot augmented by peristaltic motion to enhance locomotion, and SWTac, a novel event‑based visuotactile sensor with an actively vibrated elastomer. The event camera is mechanically decoupled from vibrations by a spring isolation mechanism, enabling high-quality tactile imaging of both dynamic and stationary objects. For algorithm design, we propose an IMU-guided temporal filter to enhance imaging consistency, improving MSNR by 24\%. Moreover, we systematically optimize SWTac with vibration parameters, event camera settings and elastomer properties. Motivated by asymmetric edge features, we also implement contact surface estimation by U-Net. Experimental validation demonstrates SWTac’s 0.2 mm texture resolution, 98\% stone classification accuracy, and 0.15 N force estimation error, while SandWorm demonstrates versatile locomotion (up to 12.5 mm/s) in challenging terrains, successfully executes pipeline dredging and subsurface exploration in complex granular media (observed 90\% success rate). Field experiments further confirm the system's practical performance.

\end{abstract}


\section{Introduction}

\IEEEPARstart{G}{ranular} materials, such as sand, gravel, and powder, exhibit complex behaviors characterized by particle flowability, frictional interactions, and heterogeneous porosity~\cite{Kou2017, gra,Katterfeld2017}. These materials are commonly encountered in underground infrastructure inspection~\cite{atta}, mineral exploration~\cite{Pal2020}, disaster rescue~\cite{REDDY2015457}, and planetary missions~\cite{vaquero_eels_2024}. In such applications, robotic systems often employ active techniques including drilling and vibration to facilitate penetration~\cite{patel2021diggerfinger,XIAO2018160, PachecoVazquez2010, Kang2018}. Due to the low visibility and unpredictable nature of these environments, tactile feedback is critical. It provides real-time information on contact forces and material properties, thereby enabling effective perception and adaptive control~\cite{10682561,5339133}.

\begin{figure}[t]
    \centering
    \includegraphics[width=\linewidth]{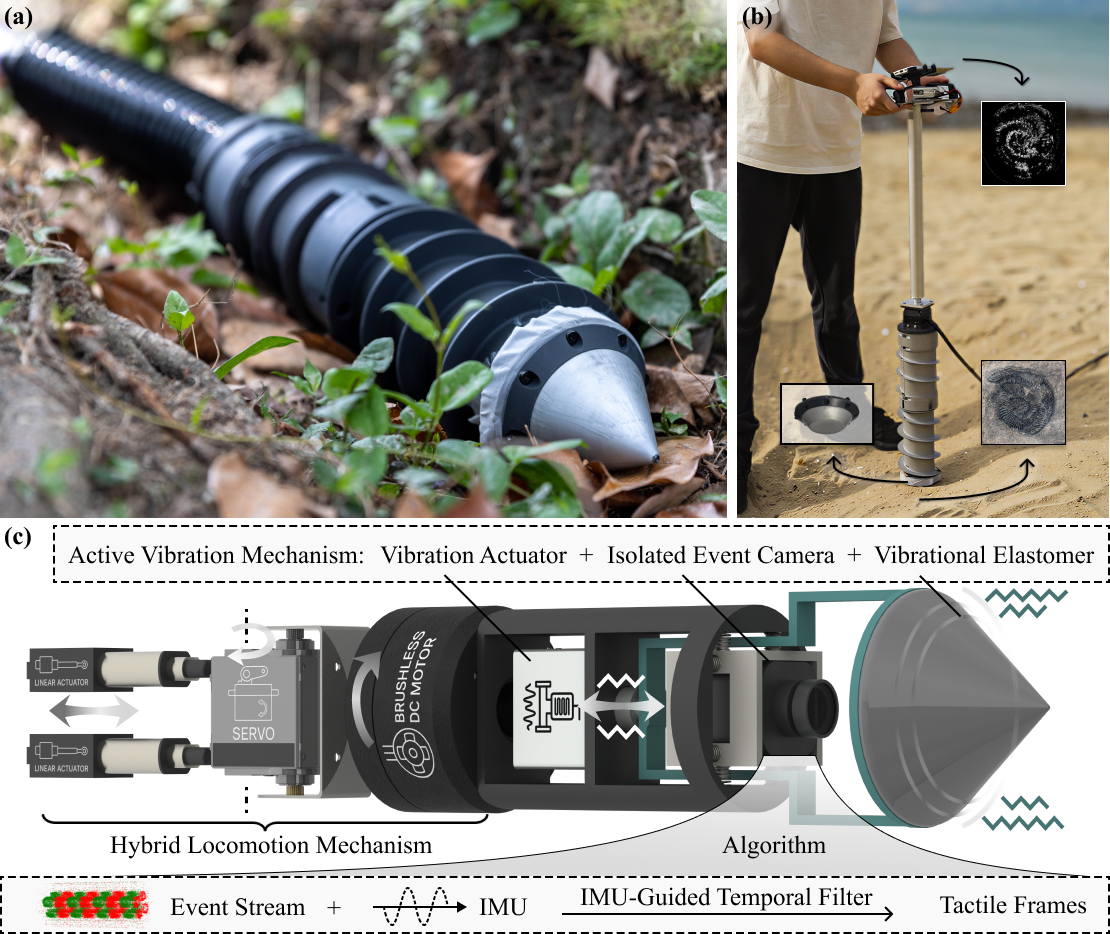}
    \caption{Overview of SandWorm. (a) The snakelike robot is navigating an outdoor trench with the conical elastomer. (b) Manually operated drilling with the frustoconical elastomer on a sandy beach, showing the texture of an ammonoid fossil.  (c) Schematic of SandWorm, featuring an integrated active vibration mechanism and hybrid locomotion system. An IMU-guided temporal filter is introduced to enhance imaging quality.
}

    \label{fig:1}
\end{figure}

Conventional electronic tactile sensors (e.g., strain gauges, piezoresistive arrays) would suffer performance degradation due to strong vibrational resonance in the drilling process~\cite{GHASEMLOONIA2015150,SALDIVAR20145169}. 
Visuotactile sensors reduce these mechanical drawbacks by separating the electronics from their sensing skin~\cite{10563188}.
However, their reliance on the camera introduces new limitations, as high-frequency vibration would induce motion blur, image trailing, and feature‑tracking errors~\cite{DASH20141634,8953329}. With the event-based imaging mechanism, event cameras can capture environmental changes in extremely short time intervals, effectively avoiding the blurring issues associated with traditional cameras due to long exposure times\cite{gallego_event-based_2022}. Moreover, with higher dynamic range, event cameras can detect subtle changes, making them ideal for precise visuotactile perception in granular media under vibration. However, event cameras intrinsically respond only to dynamic changes. To achieve static imaging, an imaging medium should be employed to generate detectable events~\cite{he_microsaccade-inspired_2024}.

Inspired by benthic polychaetes such as \textit{alitta virens}, which use sensory-rich heads and muscle-driven peristalsis combined with spiral locomotion to traverse granular sediments~\cite{sw1,sw2}, we present SandWorm, a biomimetic robotic system. SandWorm mimics the worm’s locomotion to explore granular media and confined environments, and integrates SWTac, an actively vibrated event-based visuotactile sensor for tactile perception. 
The main contributions are as follows:

\begin{itemize}

\item \textbf{Active Vibration for Static Tactile Imaging:}  
We introduce a vibration-isolation mechanism in the SWTac visuotactile sensor that generates controlled elastomer vibration, enabling static tactile perception by the mechanically isolated event camera. This approach not only achieves pixel-level tactile imaging at 1000 Hz by overcoming the inherent limitation of event cameras in capturing stationary objects, but also leverages vibration-induced granular fluidization to reduce subsurface penetration resistance.

\item \textbf{IMU-Guided Perception Algorithm :} 
To enhance event stream consistency, we propose an IMU-guided temporal filter based on a vibration-imaging model, achieving 24\% improvement in masked signal-to-noise ratio (MSNR). Furthermore, leveraging asymmetric edge features verified by finite element analysis, we employ a U-Net model for contact surface estimation, achieving a structural similarity index measure (SSIM) about 0.97. We also systematically characterize the effects of vibration frequency, amplitude, direction, event camera sensitivity, and elastomer viscoelasticity in SWTac.

\begin{table}[b]
    \centering
    \caption{Comparison of invasive robotic perception methods }
    \renewcommand{\arraystretch}{1.2} % 增大行间距
    \label{tab:compare_sand}
    \resizebox{\columnwidth}{!}{
    \begin{tabular}{l l l c}
        \toprule
        \textbf{Reference} & \textbf{Sensor} & \textbf{Resolution} & \textbf{Force \& Contact}\\
        \midrule
        Richter et al.~\cite{richter2022arcsnake}
            & Camera
            & Pixel Level, $\sim$30 Hz
            & \xmark \\
        Xue et al.~\cite{xue2023contact}
            & Camera
            & Pixel Level, $\sim$30 Hz
            & \xmark \\
        Kwon et al.~\cite{kwon2012pipeline}
            & Camera
            & Pixel Level, $\sim$30 Hz
            & \xmark \\
        Kakogawa et al.~\cite{kakogawa2016screw}
            & Camera
            & Pixel Level, $\sim$30 Hz
            & \xmark \\
        Nourizadeh et al.~\cite{nourizadeh2024skid}
            & IMU
            & Single Point, 5 Hz
            & \xmark \\
        Kolvenbach et al.~\cite{kolvenbach2019haptic}
            & Force
            & Single Point, 2 Hz
            & \cmark \\
        Ruppert et al.~\cite{ruppert2020foottile}
            & Force
            & Single Point, 330 Hz
            & \cmark \\
        Russell et al.~\cite{russell2005mole}
            & Chemistry
            & Single Point, 0.003 Hz
            & \xmark \\
        Digger Finger~\cite{patel2021diggerfinger}
            & Visuotactile
            & Pixel Level, $\sim$30 Hz
            & \cmark \\
        \midrule
        SandWorm (Ours)
            & Visuotactile
            & \textbf{Pixel Level, 1000 Hz}
            & \textbf{\cmark} \\
        \bottomrule
    \end{tabular}
    }
\end{table}

\begin{table*}[bp] % 使用table*环境
    \centering
    \caption{Comparison of event-based sensors for robotic applications}
    \renewcommand{\arraystretch}{1.2} % 增大行间距
    \label{tab:compare_event}
    \resizebox{2\columnwidth}{!}{
\begin{tabular}{llllllll}
\toprule
\textbf{Sensor}   & \textbf{Latency}            & \textbf{Spatial Resolution}   & \textbf{Force Sensing}    & \textbf{Sensing Type} & \textbf{Camera} & \textbf{Texture} & \textbf{Characteristics} \\
\midrule
Noise2Image \cite{Cao:25}        & 100 ms    &  N/A  &  N/A &  Vision  & EV (Dynamic + Static)  &   N/A   &  Static Imaging for EV, Software \\
Yousefzadeh et al.~\cite{active}    & 100 ms             & 1 mm                &\xmark          & Vision      & EV (Dynamic + Static)             &  \cmark               &  Static Imaging for EV, Hardware          \\
AMI-EV \cite{he_microsaccade-inspired_2024}   & \textbf{0.8 ms}          & \textbf{0.04 mm }               & \xmark      & Vision    &    EV (Dynamic + Static)     &        \cmark       &    Static Imaging for EV, Hardware   \\
NeuTouch \cite{taunyazov_event-driven_2020}  & 10 ms          & 1 mm              & \cmark    & Vision, Tactile  & RGB + EV (Dynamic)         &      \cmark    &     Multimodal Processing   \\
Baghaei et al.~\cite{BaghaeiNaeini2020}       & 21 ms             & 0.2 mm                     & 0.16 N           & Vision, Tactile   & EV (Dynamic)             & \cmark            &  Force Estimation                      \\
GelEvent \cite{yin_geleventnovel_2025}       & 5.5 ms         & 225 markers (3 mm)             &  0.8 N  & Visuotactile &   EV (Dynamic)      &      \xmark            &   Contact Area Estimation           \\
E-BTS \cite{mukashev_e-bts_2025}             & 2 ms            & 5 markers               &  Shear 0.27 N & Visuotactile  &  EV (Dynamic)       &          \xmark        &     Teleoperation    \\
Evetac \cite{funk_evetac_2024}               & 1 ms         & 63 markers               & Shear 0.22 N  & Visuotactile  & EV (Dynamic)    &       \xmark        &      Slip Detection      \\
Digger Finger~\cite{patel2021diggerfinger}   &  33 ms            & 0.3 mm          & \cmark          & Visuotactile  & RGB (Static)   &      \cmark        & GelSight~\cite{Yuan2017GelSight} in Granular Media \\
\bottomrule
SWTac (\textbf{Ours})  & \textbf{1 ms}    & \textbf{0.2 mm}    & \textbf{Shear 0.15 N}    & Visuotactile    & EV (Dynamic + Static) & \cmark & \makecell[l]{Granular Media\\Static Imaging for EV, Hardware} \\

\bottomrule

    \end{tabular}%
    }
\captionsetup{justification=centering}
\end{table*}

\item \textbf{Hybrid Locomotion Mechanism:} We introduce a hybrid locomotion mechanism that synergistically combines screw-driven rotation with pushrod-actuated peristalsis for enhanced propulsion in confined environments. Implemented in SandWorm, a snakelike robot integrated with SWTac, this mechanism improves propulsion efficiency while enabling adaptable application.

\item \textbf{Real-World Field Validation:}  
Extensive experiments demonstrate SWTac's capabilities in precise force estimation (0.15 N accuracy), high-resolution texture sensing (0.2 mm resolution), and robust stone classification (98\% accuracy). SandWorm demonstrates robust performance across diverse field scenarios, successfully executing high-curvature movements, steering and multi-terrain traversal (dense grass, tangled bushes, cement surfaces) at speeds up to 12.5 mm/s; performing complex tasks such as pipeline dredging and subsurface exploration in non-uniform media (beach sand, garden soil, four industrial granular media); and achieving an observed 90\% success rate in detecting subsurface objects.

\end{itemize}

The remainder of this paper is organized as follows. Section II reviews related works. Section III describes the design and framework of SWTac and SandWorm. Section IV presents the algorithms for enhancing event-based imaging, and characterization of SWTac. Section V details the experimental evaluations. Section VI discusses insights and future work. Finally, Section VII concludes this article.

% \subsection{Non-Invasive Perception in Granular Media}

\section{Related Works}
\subsection{Perception in Granular Media}

Perception in granular media poses unique challenges that have motivated a variety of non‐invasive and invasive solutions. Current non-invasive methods, such as LiDAR~\cite{ZAHIRI2021102603}, ultrasonic wave propagation~\cite{GHEIBI2018112}, and Rayleigh waves~\cite{Lee} struggle in granular environments due to factors like clutter, material heterogeneity, and deformation under stress. Techniques like MRI~\cite{Magnetic} and dynamic X-ray radiography~\cite{Guillard2017}, are useful for inferring media properties rather than detecting subsurface objects.

% Similarly, the GRAINS system~\cite{zhang2023grainsproximitysensingobjects} and Stepped Frequency Continuous Wave (SFCW) ground penetrating radar~\cite{10939276} rely on specific granular phenomena and signal-to-noise ratio (SNR) issues, which limits their effectiveness for complex subsurface exploration.

Invasive sensing is therefore essential for subsurface exploration. Table~\ref{tab:compare_sand} summarizes existing invasive robotic perception approaches. Robots operating in granular media~\cite{xue2023contact,richter2022arcsnake} and confined environments~\cite{kakogawa2016screw,kwon2012pipeline} typically rely on cameras for perception, but these systems can only capture surface features and fail to image once the camera is in contact with the medium due to blocked illumination. Specialized invasive sensors are designed with penetration ability, but may suffer from prohibitively long detection times~\cite{russell2005mole}. Force sensor~\cite{kolvenbach2019haptic,ruppert2020foottile} and IMU~\cite{nourizadeh2024skid} are commonly used for locomotion control, yet they measure only localized contact information and cannot provide the broad coverage like vision-based methods. In contrast, visuotactile sensors integrate illumination, imaging, and contact skin into a single module~\cite{10563188}, making them ideal for exploration in granular media. For example, DiggerFinger~\cite{patel2021diggerfinger} performs exploration by GelSight~\cite{Yuan2017GelSight} in a sandy environment, but its imaging performance would degrade under the strong vibrations required for medium penetration.

% For locomotion, IMUs~\cite{nourizadeh2024skid,richter2022arcsnake} are commonly employed. To facilitate interaction with external environments, robots often use force sensors~\cite{kolvenbach2019haptic,ruppert2020foottile,zhu2023simulation}. However, external observation becomes impractical in subterranean or pipeline environments, where visual quality is constrained by lighting conditions.
% In these scenarios, visuotactile sensors~\cite{patel2021diggerfinger,Yuan2017GelSight} offer superior stability, providing greater sensing area, resolution, and force detection range. Particularly, event cameras are better suited to addressing the motion blur that traditional RGB cameras experience in high-vibration environments, such as those encountered when exploring in granular media. 

% But in high-vibration environments, such as exploring in granular media, traditional RGB cameras suffer from  and poor performance in low-light conditions.

\subsection{Visuotactile Sensors with Event Camera}

% In contrast, event cameras could capture dynamic changes with high temporal resolution, avoiding motion blur and maintaining clarity even at high vibration frequencies. However, event cameras could capture only dynamic changes due to its intrinsic limitation. 

Event cameras effectively mitigate motion blur, which is a common issue for traditional RGB cameras in dynamic environments~\cite{8946715,gallego_event-based_2022}, such as those arising from vibration-driven exploration.
As shown in Table~\ref{tab:compare_event}, existing event-based visuotactile sensors generally based on markers~\cite{funk_evetac_2024,mukashev_e-bts_2025,yin_geleventnovel_2025}. They follow the design of conventional visuotactile sensors (e.g. GelSight~\cite{Yuan2017GelSight}), only replacing the RGB camera with event camera. 
By employing advanced elastomer designs, they achieve higher spatial resolution for tasks including force and contact surface estimation~\cite{yin_geleventnovel_2025}. 
Although these sensors are particularly useful in teleoperation~\cite{mukashev_e-bts_2025} and sliding detection~\cite{funk_evetac_2024}, their reliance on markers obstructs the field of view and limits imaging to taxel‐level resolution rather than pixel‐level~\cite{patel2021diggerfinger}, resulting in poor texture perception and therefore unsuitable for exploration tasks. 

\begin{figure*}[ht]
    \centering
    \includegraphics[width=0.98\linewidth]{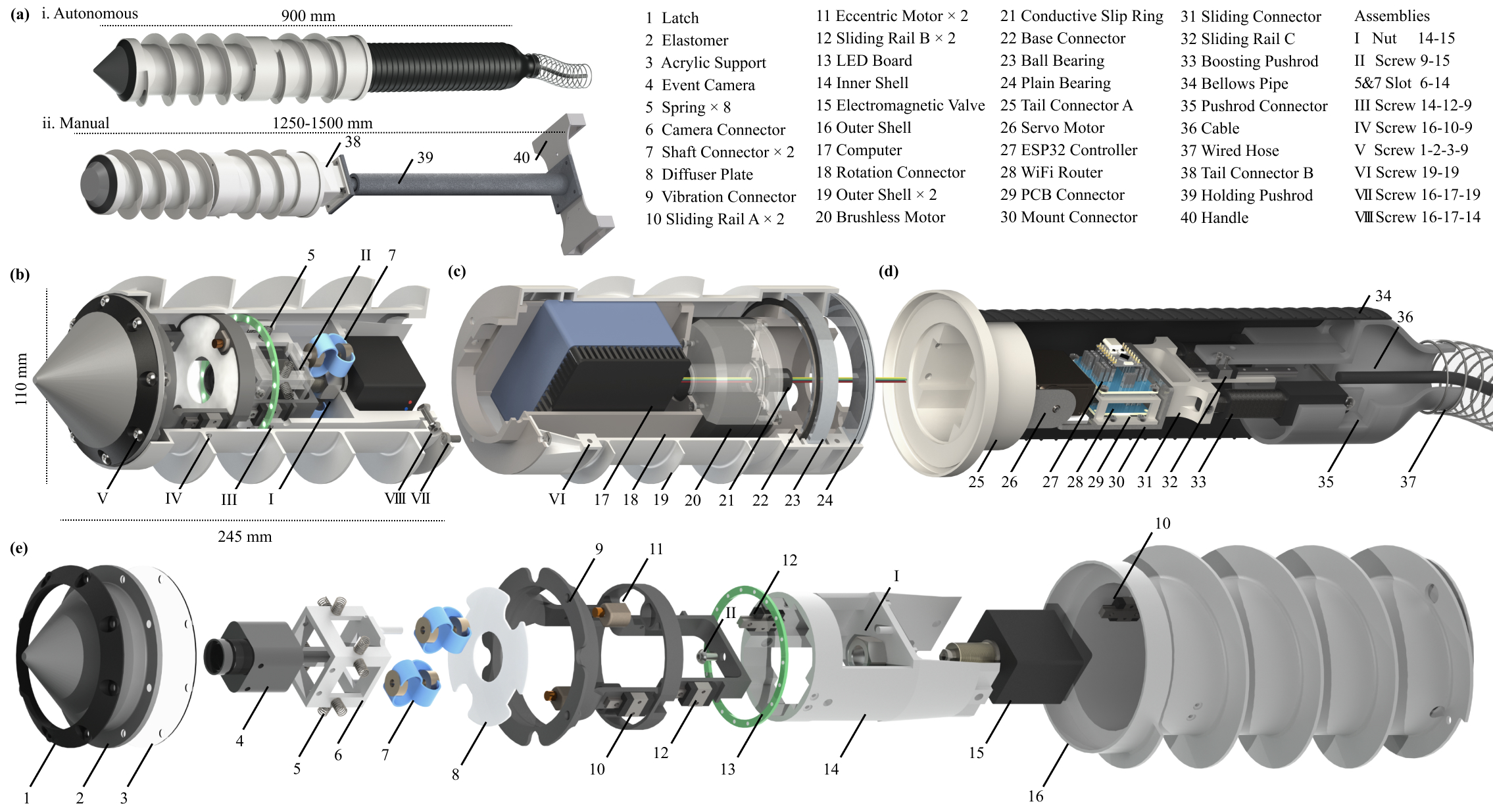}
    \caption{Overview of SandWorm's mechanical structure. (a) Rendered views of SandWorm: an autonomous snakelike robot and a manually operated device. Cross-sectional views of (b) SWTac, the perception module, (c) the rotational driving module and (d) the snakelike tail module. (e) Exploded view of SWTac.}
    \label{fig:structure}
\end{figure*}

If not relying on markers, existing methods fix the event camera externally to the tactile sensor to capture the dynamic deformations of the sensing skin during contact~\cite{taunyazov_event-driven_2020,BaghaeiNaeini2020}. Although effective for texture extraction, these approaches produce experimental setups to make the event camera remaining stationary, rather than proposing a compact, integrated visuotactile sensor. The primary challenge lies in the fact that event cameras only capture dynamic changes~\cite{gallego_event-based_2022}. In integrated visuotactile systems, the sensing skin and event camera remain stationary relative to each other, so events are generated only during transient deformations~\cite{yin_geleventnovel_2025}, resulting in sparse and inconsistent imaging. Markers have predetermined positions, enabling localization of taxels despite these gaps. But in visuotactile sensors without markers, the absence of predictable features makes it difficult to acquire high‐quality and consistent event‐based tactile images.

Several vision-based solutions have been proposed to address the intrinsic limitation on static imaging of event cameras. Software-based noise sampling recovery methods~\cite{Cao:25} demand high processing time. For hardware-based solutions, directly vibrating the event camera carries the risk of mechanical damage~\cite{active}. A better approach is to add an imaging medium outside the sensor to trigger events, such as using a rotating prism to redirect the light path while the camera remains static~\cite{he_microsaccade-inspired_2024}. Particularly, in visuotactile sensors, the elastomer could serve as a natural medium between the event camera and the object. Therefore, without requiring additional imaging medium, introducing active vibration to the elastomer would enable the event‐based visuotactile sensor to capture a high‐quality and consistent tactile event stream.

In this article, we present SandWorm, a robot designed for navigation and exploration in both granular media and confined environments (Fig.~\ref{fig:structure}(a)). SandWorm comprises two main modules: an actuation module featuring a screw-actuated body enhanced with peristaltic motion for improved mobility, and a perception module, SWTac. SWTac is an active event-based visuotactile sensor that applies controlled vibration to the elastomer while mechanically isolating the event camera (Fig.~\ref{fig:structure}(b)). This active vibration effectively integrates perception and actuation, as it guarantees consistent tactile perception and simultaneously facilitates invasive exploration by loosening the granular material.

\section{Hardware Structure}

This section first details the hardware structure of SWTac, including the vibration-isolation mechanism, elastomer, and illumination design. We then introduce the actuation module, followed by an analysis of the rotational–peristaltic locomotion mechanics. Finally, we summarize SandWorm's perception-actuation framework to highlight our core innovative modules and their interconnections.

\subsection{Perception Module Design}\label{sec:swtac}

The structure of SWTac is designed to efficiently transmit active vibrations to the elastomer while simultaneously isolating the event camera. Fig.~\ref{fig:structure}(b) shows the cross-sectional view of SWTac, and Fig.~\ref{fig:structure}(e) presents a detailed schematic breakdown of its components.  The core of this visuotactile sensor is the DVXplorer Mini event camera, selected for its compact size. The peripheral structure is designed around this camera and features two functionally distinct sections: a vibrational part and an isolated part.

For the vibrational part, the design aims to generate and guide multi-axis vibration using a compact assembly. This is achieved by mounting both the primary vertical actuator (electromagnetic valve, P15) and the supplementary horizontal actuators (eccentric motors, P11) onto a shared vibration connector (P9). This entire dynamic assembly is then coupled to the stationary inner shell (P14) via sliding rails (P10, P12). These rails function to constrain the motion primarily to the vertical axis while accommodating assembly tolerances, ensuring stable operation.

For the isolated part, the core design is to mechanically decouple the event camera (P4) from the high-vibration assembly. This is accomplished by soft-coupling the camera (mounted on P6) to the stationary inner shell using a dedicated multi-axis isolation system: eight springs (P5) provide horizontal isolation, and two flexible shaft connectors (P7) provide vertical isolation. This design effectively protects the sensitive electronics without obstructing the camera's field of view. Finally, the LED board (P13) and diffuser plate (P8) are integrated for illumination. The whole assembly is secured by the primary load-bearing (Assembly VIII) and enclosed within the spiral outer shell (P16) to enhance drilling performance.

The elastomer serves as a fundamental component in visuotactile sensors. In granular media, elevated abrasive forces necessitate a thicker elastomer layer than those used in conventional designs to enhance durability. Therefore, polydimethylsiloxane (PDMS; Sylgard™ 184, Dow Corning) is adopted to guarantee optical transparency. The assembled elastomer tip (P1-P3) is installed at the end of the vibration connector. The vibration-isolation structure enables dual-axis vibrational control of the elastomer tip, both horizontally and vertically.

\begin{figure}[t]
    \centering
    \includegraphics[width=\linewidth]{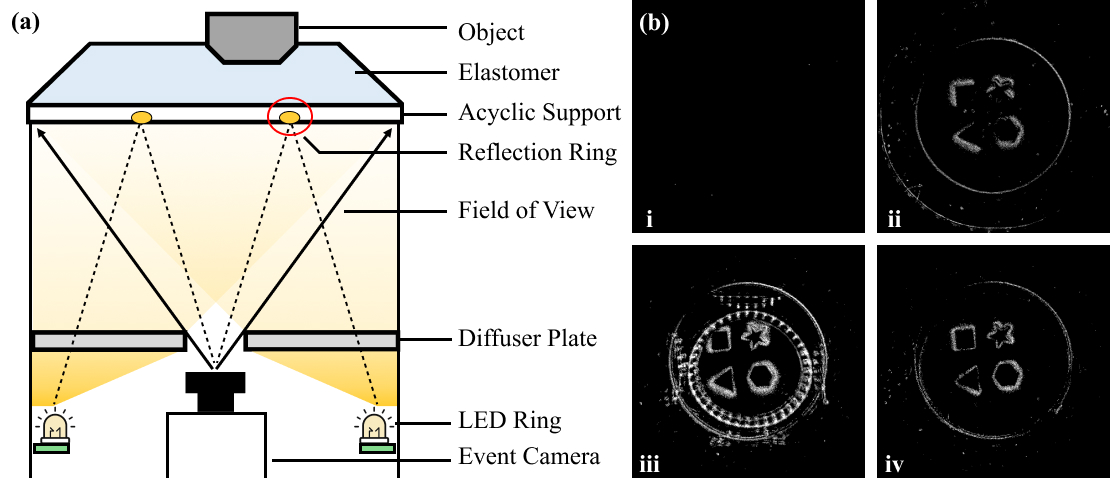}
    \caption{Illumination setup and conditions.
(a) Hardware setup and optical‑path layout. 
(b) Reconstructed images of event camera with
i. No illumination;
ii. Direct dark‐field illumination;
iii. Direct bright‐field illumination;
iv. Diffuse bright‐field illumination with a diffuser plate.}
    \label{fig:illu}
\end{figure}

The illumination system is crucial to event cameras, as sufficient lighting is essential for enhancing imaging quality. In our design, we utilize a circular array of 14 LEDs (WS2812B 5V) with a diffuser plate.
As illustrated in Fig.~\ref{fig:illu}, with direct lighting, a peripheral reflective ring is observed. Direct dark‐field illumination would reduce the quality of the primary signal by diverting the light path from a direct frontal configuration to a lateral one.
Fig.~\ref{fig:illu}(b)iv shows the capture result by diffuse bright‐field illumination. The diffuser plate homogenizes the light distribution while attenuating its intensity, thereby reducing the severity of specular reflections. Additionally, the central opening of the diffuser plate is also optimized to provide a hardware feature mask that delineates the region of interest for subsequent software processing.

\subsection{Actuation Module Design}\label{sec:sandworm}

Snakelike robots are commonly used for exploration in granular environments. These robots utilize different actuating methods, including lateral wave motion~\cite{branyan2018soft}, peristaltic motion~\cite{scheraga2020peristaltic,chen_bioinspired_2024} and screw mechanisms~\cite{richter2022arcsnake,vaquero_eels_2024}. Screw mechanisms provide both propulsion and steering forces, making them particularly effective for navigating in confined spaces~\cite{dachlika_mechanics_2020} and pipelines~\cite{richter2022arcsnake}. Therefore, we adopt a rotational‐peristaltic driving scheme for locomotion, which requires fewer actuators compared to lateral wave methods and enhances traction in granular media. As depicted in Fig.~\ref{fig:structure}(a), SandWorm has a reconfigurable design, and can work either as a autonomous snakelike robot or as a manually operated device. Both configurations differ only at the tail end, while the perception module (SWTac, Fig.~\ref{fig:structure}(b)) and the rotational driving module (Fig.~\ref{fig:structure}(c)) are identical.

The rotational driving module (Fig.~\ref{fig:structure}(c)) provides both data processing and screw actuation. An onboard computer (P17) executes event–stream processing and high-level controlling. A brushless motor (P20) is employed to continuously rotating the SWTac sensor. The drivetrain incorporates a coaxial bearing stack and a dedicated mechanical interface, ensuring clean separation between the rotating and non-rotating modules. A spiral outer shell surrounds the assembly, implemented as two semi-circular housings (P19, Assembly VI) that enclose the periphery of the module. The shell is machined from an Al–Mg alloy to improve structural robustness and wear resistance during contact with granular media.

The tail module (Fig.~\ref{fig:structure}(d)) hosts the remaining actuation. It embeds electronics within a collapsible bellows conduit (P34). Steering actuation is provided by a servo motor (P26), while peristaltic motion is generated by a pushrod mechanism (P33). An ESP32 microcontroller (P27) and a compact router (P28) serve as the low-level controller and wireless communication backbone. Linear guides (P32) and reinforced metal connector (P30) increase stiffness under load. At the distal end, a compliant, non-extendable support hose (P37) with cable (P36) inside is employed. For extended applications, a pushrod (P39) of larger size will be mounted to the rotational driving module, forming a hand‑held (P40) drilling assembly.

\begin{figure}[t]
    \centering
    \includegraphics[width=\linewidth]{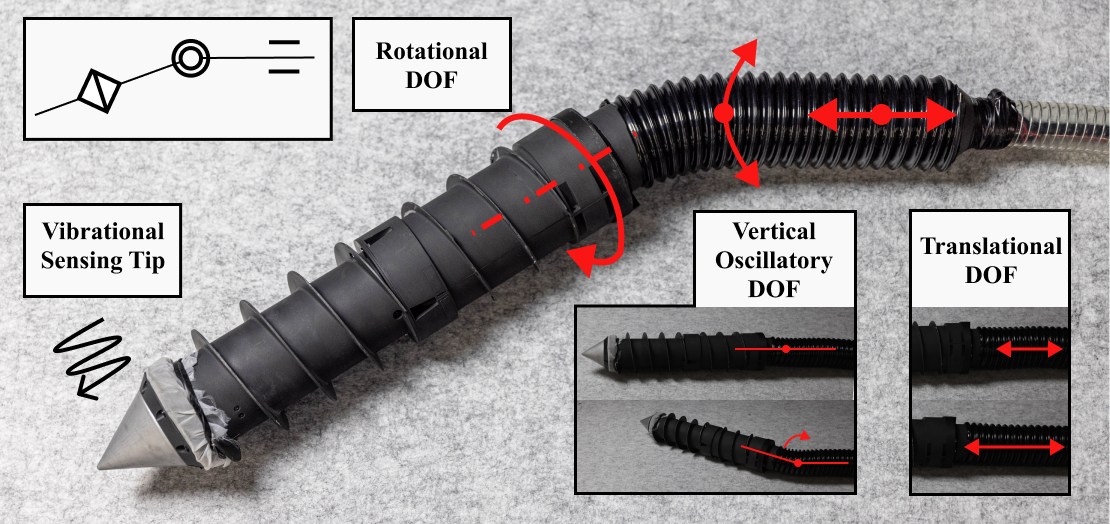}
    \caption{Overview of the SandWorm robot's DOFs. Locomotion is achieved by combining rotation, peristaltic translation, and vertical oscillation. The vibrational elastomer tip for perception is also illustrated.}
    \label{fig:kinematic}
\end{figure}

\subsection{Locomotion Mechanics}
\label{sec:kinematics}

SandWorm provides three degrees of freedom (DOFs) of locomotion, as shown in Fig.~\ref{fig:kinematic}: a rotation of the spiral shell driven by a brushless motor; a servo-driven vertical oscillation for steering; and a peristalsis motion via internal pushrods for linear translation. The front rotational axis is rigidly coupled to SWTac enclosed by the spiral shell. The rear of SandWorm is guided by a flexible hose which supplies reaction force. Consequently, with its tail serving as a pivot, SandWorm leverages frictional contact to realize forward motion.

To realize robust locomotion under frictional contact, SandWorm employs a peristaltic drive made of two alternating phases that combine screw traction with pushrod actuation. In confined environments (e.g., pipelines), gravity together with the pipe support provides roll stability so that a single-chirality screw can generate forward thrust without requiring paired counter-rotating screws. Let $F_{\text{p}}$ denote the axial force applied by the pushrod and $F_{\text{friction}}$ the lumped friction opposing motion. The effective gravitationally resolved driving component along the axis is denoted $F_{\text{propel}}$ (cf. Appendix~\ref{app:kinematic-derivations}). During the extension phase, the pushrod assists propulsion, leading to the net force:
\begin{equation}
F_{\text{extension}} = F_{\text{propel}} + F_\text{p} - F_{\text{friction}},
\label{eq:net-extension}
\end{equation}
which produces a rapid forward stroke. During retraction, the pushrod opposes motion, and the net force becomes:
\begin{equation}
F_{\text{retraction}} = F_{\text{propel}} - F_\text{p} - F_{\text{friction}},
\label{eq:net-retraction}
\end{equation}
yielding little displacement. This asymmetric two-phase actuation, akin to a controlled shift of the robot’s center of mass, produces a net advance greater than that of continuous rotation alone.

\subsection{Perception-Actuation Framework}

\begin{figure}[t]
    \centering
    \includegraphics[width=\linewidth]{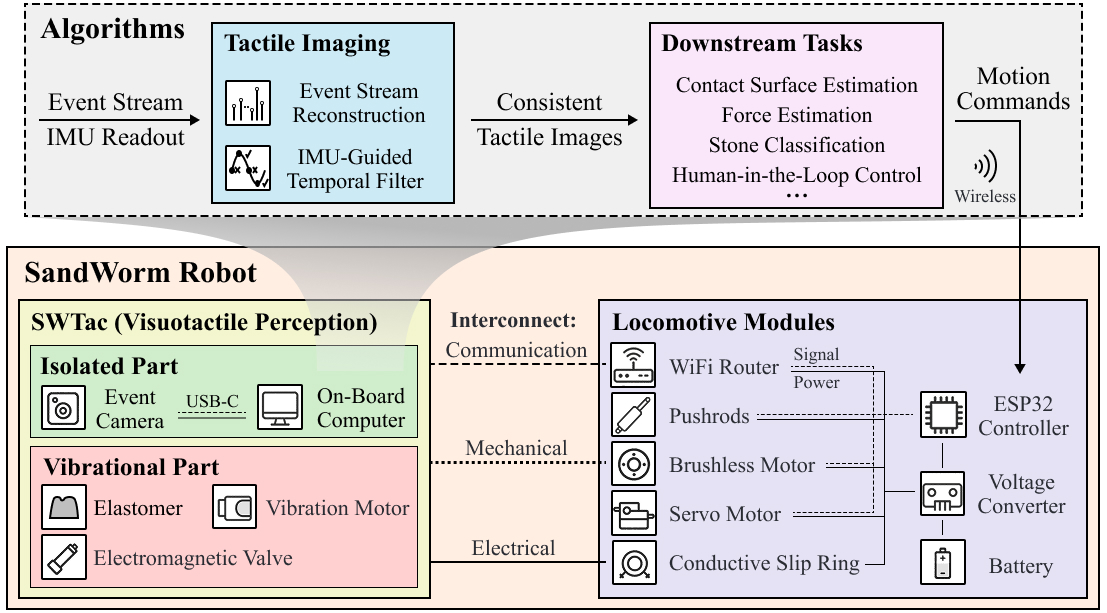}
    \caption{System architecture of the SandWorm robot, detailing the hardware and algorithm integration. The hardware (bottom) features the SWTac visuotactile sensor (comprising an isolated and a vibrational part) and the locomotive modules. The algorithmic pipeline (top) processes the event stream and IMU readout from the event camera, including an IMU-guided temporal filter followed by downstream tasks.}
    \label{fig:control}
\end{figure}

The system architecture of SandWorm, as depicted in Fig.~\ref{fig:control}, integrates the SWTac visuotactile sensor with locomotive modules, as well as a hierarchical signal processing pipeline. The perception component is the SWTac sensor. Its vibrational part includes the actuators, while its isolated part contains the event camera and the on-board computer. The robot's actuation is handled by the locomotive modules, comprising pushrods, a brushless motor, and a servo motor, all managed by an ESP32 controller and a central battery. The interconnection between the SWTac and the locomotive modules is threefold. Brushless motor mechanically couples the sensor to the locomotive body. Conductive slip ring delivers power from the non-rotating locomotive section to the rotating SWTac as the electrical link. The communication link connects the on-board computer to the ESP32 by WiFi.

The algorithmic pipeline (top panel) runs entirely on the SWTac's on-board computer. Raw sensor data, specifically the event stream and IMU readout, are used to reconstruct frames and are processed by our IMU-guided temporal filter, which generates consistent tactile images. This processed data is then passed to the downstream tasks (e.g., contact surface estimation, force estimation) to generate high-level motion commands, either directly or indirectly through human-in-the-loop control (utilizing remote real-time image viewing). These commands are transmitted wirelessly via the communication interconnect to the ESP32 controller in the locomotive modules, which in turn actuate the motors and pushrods. Thus, the SWTac's tactile perception, which relies on vibration and motion, acts back on the actuators to form a fundamental perception-actuation loop.

\section{Algorithm Design}

% Based on the hardware design, SWTac is enabled to effectively detect stationary objects using event camera.
In this section, we first introduce the workflow to reconstruct grayscale images from the asynchronous event stream. To mitigate vibration‐induced temporal inconsistencies, we develop an IMU‑guided temporal filter based on the theoretical model of event‐based imaging under vibration.
After that, motivated by the asymmetric edge features in reconstructed tactile images, we conduct contact surface estimation, beyond the edge information captured by the event camera.
Finally, we characterize the imaging performance of SWTac by different parameters, including vibration amplitude, frequency, direction, event camera sensitivity and elastomer viscoelasticity.

\subsection{Event-based Image Reconstruction}
\label{sec:calib}

\begin{figure}[t]
    \centering
    \includegraphics[width=\linewidth]{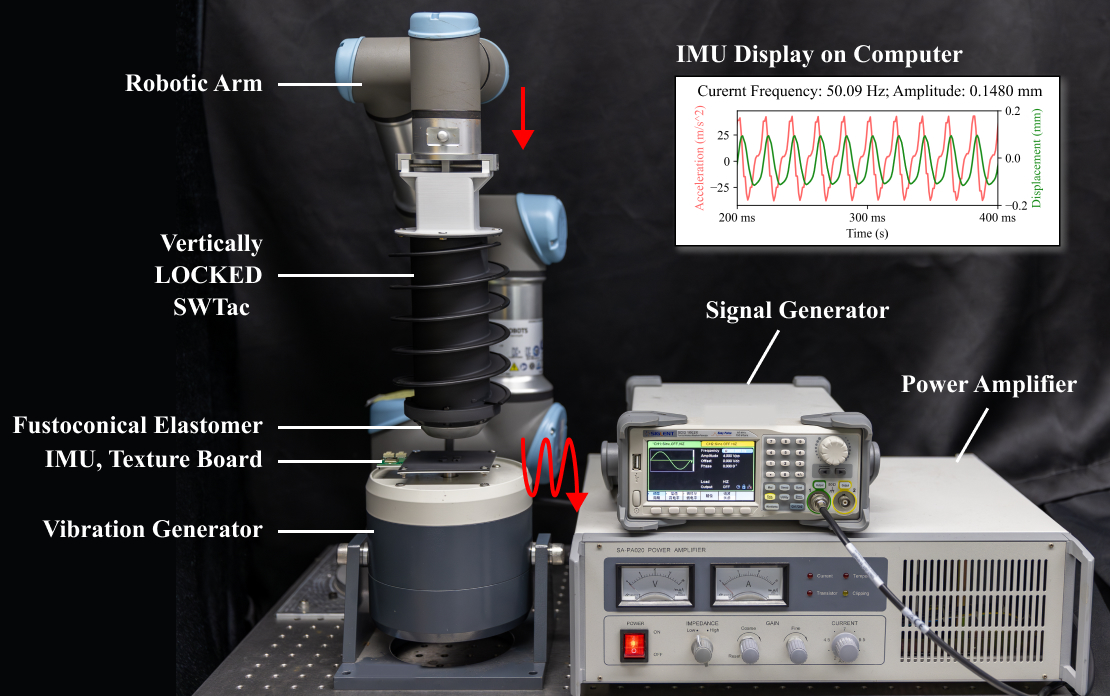}
    \caption{Data acquisition platform.
    SWTac with frustoconical elastomer is connected to the robotic arm.
The texture board and IMU are attached to a programmable vibration generator.}
    \label{fig:calib}
\end{figure}

\subsubsection{\textbf{Hardware Setup}}
To enhance perception performance by algorithm design, we set up a data acquisition platform in Fig.~\ref{fig:calib}. Instead of using the electromagnetic valve, we utilized a vibration generator to provide active vibration. This generator is driven by a signal generator with a power amplifier, providing explicit frequency control and implicit amplitude control via adjustable power of stable sine wave. An IMU mounted on the generator communicates with a computer in real time to compute the amplitude, enabling adjustments. A UR5 robotic arm was employed to position the sensor accurately above the vibration generator.

To ensure the effect of the external vibration generator is equivalent to the sensor's active vibration, we modified the sensor housing by removing the vertical elastic DOF, thereby rigidly securing the elastomer. Externally induced vibrations technically would produce the same effect as actively vibrating the elastomer, given that the sensor remains stationary in both setups. We verified the interchangeability of these two actuators in Section~\ref{sec:exphard}. The frustoconical elastomer with a flat surface was used for data collection. 

For the event camera settings, we used three threshold levels of the camera change-detection circuitry to assess sensitivity~\cite{4444573}. The threshold defines the minimal log intensity variation at the pixel level required to induce an event. Lower thresholds enhance sensitivity by generating more events in response to subtle light changes, while higher thresholds diminish sensitivity by excluding minor intensity variations. We employed these three threshold values to capture event streams corresponding to different camera sensitivity settings.

\begin{figure}[t]
    \centering
    \includegraphics[width=\linewidth]{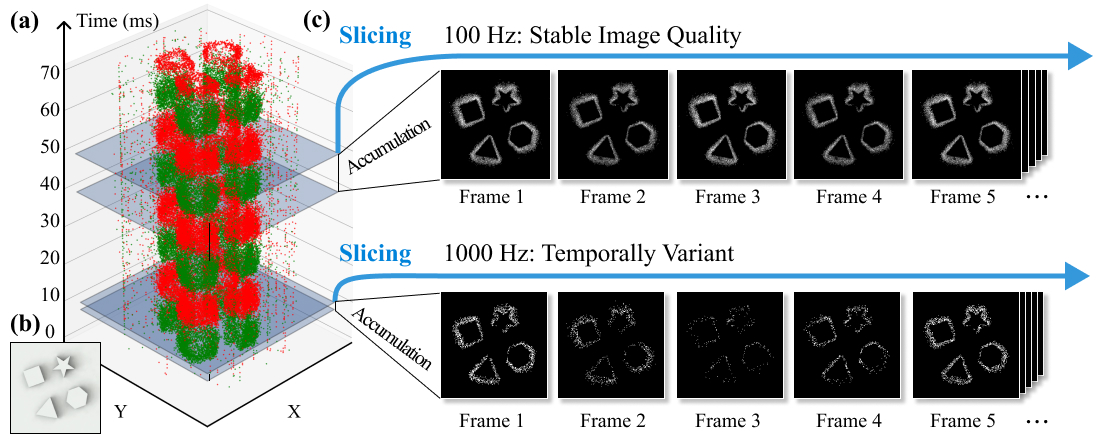}
    \caption{Principle and results of grayscale image reconstruction.
(a) Event stream during vibration, with positive events in red and negative events in green.
(b) Texture board.
(c) Grayscale images under different reconstruction frame rates.}
    \label{fig:gray}
\end{figure}

\subsubsection{\textbf{Reconstruction Workflow}}
The captured event streams are sequences of discrete events $\mathcal{E}=\{e_i\}$, each encoding $(x_i, y_i, p_i, t_i)$ as a pixel coordinate, a brightness-change polarity  and a precise timestamp. An example event stream is depicted in Fig.~\ref{fig:gray}(a). Under the active vibration of 50 Hz in SWTac, the temporal sequence of the event stream shows an alternating pattern with color-encoded polarity.

Next, the event stream is processed to a sequence of frame-based grayscale images $G_k$, as shown in Fig.~\ref{fig:gray}(c). This conversion is achieved by event accumulation. As a pre-processing step, an active background filter was adopted to suppress noise from the raw stream, yielding a filtered event set $\mathcal{E}_{\text{filtered}}$. Subsequently, we convert these asynchronous events into frames at $f_{rate}$ frame rate by slicing the filtered stream into discrete time windows of a specific accumulation time, $\Delta T_{\text{acc}}=1/f_{rate}$, and projecting the events within each slice onto the X-Y plane. Each event contributes a fixed numeric value $C$, namely the contribution, to its pixel’s brightness. This process is formally defined as:
\begin{equation}
G_k(x, y) = \sum_{
\substack{
e_i \in \mathcal{E}_{\text{filtered}} \\
(x_i, y_i) = (x, y) \\
t_i \in [T_k, T_k + \Delta T_{\text{acc}})
}
} C
\label{eq:Gk}
\end{equation}
where $G_k(x, y)$ is the intensity of frame $k$ at pixel $(x, y)$, and $T_k$ is the start time of the frame. This formulation inherently applies a step-function reset between windows, preventing carry-over artifacts. Moreover, by discarding polarity information (the $p_i$ term is omitted), we rely solely on event counts per pixel, which further sharpens edges in the reconstructed frames. The data processing of event camera was primarily conducted using the DV-processing library~\cite{dv-processing}.

\begin{table}[ht]
  \centering
  \caption{Parameter settings for grayscale image reconstruction}
  \begin{tabularx}{0.95\columnwidth}{
    >{\centering\arraybackslash}X   % 第一列
    >{\centering\arraybackslash}X   % 第二列
    >{\centering\arraybackslash}X   % 第三列
  }
    \toprule
    Accumulation Time & Contribution & Reconstruction Rate \\
    \midrule
    1 ms  & 1    & 1000 Hz \\
    2 ms  & 1    & 500 Hz  \\
    5 ms  & 0.5  & 200 Hz  \\
    10 ms & 0.25 & 100 Hz  \\
    33 ms & 0.08 & 30 Hz   \\
    \bottomrule
  \end{tabularx}
  \label{tab:gray_image_params}
\end{table}

With carefully chosen parameters for accumulation time and contribution, we could achieve real‐time grayscale reconstruction at up to 1000 Hz, as listed in Table~\ref{tab:gray_image_params}. However, the alternating pattern of the event stream directly influences the reconstruction quality. Fig.~\ref{fig:gray}(c) shows that a low reconstruction rate (e.g., 100 Hz) produces consistent images that do not capture the vibration pattern. In contrast, a higher reconstruction rate (e.g., 1000 Hz) captures the dynamic variations of the vibration pattern, resulting in image inconsistency.

\subsubsection{\textbf{Imaging Quality Metrics}}
To assess the quality of the reconstructed grayscale images, we introduce MSNR, Shannon entropy and MSE. MSNR is defined as the masked signal-to-noise ratio calculated by excluding most of the uninformative black background. Specifically, we compute the SNR only within the foreground region $\Omega$ as follows:
\begin{equation}
\label{eq:MSNR}
\text{MSNR} = 10 \lg \left( \frac{\sum_{(i)\in\Omega} I(i)^2}{\sum_{(i)\in\Omega}\left|I(i)-\mu_\Omega\right|^2} \times \frac{N_\Omega}{N_{\text{Image}}} \right),
\end{equation}
where $I(i)$ is the grayscale value of the $i$-th pixel in the image, and $\mu_\Omega$ is the mean grayscale value over $\Omega$. $N_\Omega$ and $N_{\text{Image}}$ represent the number of pixels in region $\Omega$ and the image. 

The Shannon entropy $S$ is adopted to quantify the information content of the image:
\begin{equation}\label{eq:Entropy}
S = - \sum_{l=0}^{L-1} q_l \log_{2} q_l,
\end{equation}
where $q_l$ is the occurrence probability of the $l$-th grayscale level and $L$ is the number of grayscale levels.

The MSE metric is used to evaluate the similarity between the reconstructed image and the ground truth edges. To establish the groundtruth image, we first extract the edges from the corresponding texture board model, then compute the undistorted coordinate transformation by camera calibration. The MSE is calculated as:
\begin{equation}\label{eq:MSE}
\text{MSE} = \frac{1}{N_\Omega} \sum_{i=1}^{N_\Omega} \left| I(i)-I_\text{GT}(i) \right|^2,
\end{equation}
where $I_\text{GT}(i)$ is the corresponding pixel value of ground truth.

\subsection{IMU-Guided Temporal Filter}
\label{sec:imu}

In this part, we propose the IMU-guided temporal filter, addressing the temporal inconsistency of the event stream. We formulated and validated the mathematical model of event-based visuotactile perception with active vibration, proposed a deployment strategy and analyzed the statistical results.

\subsubsection{\textbf{Motivation}}

While an active vibration strategy is essential to make static objects visible to the event camera, it also introduces oscillatory interaction between the elastomer and external objects. Since event cameras are highly sensitive to these temporal variations, the vibration induces fluctuations in event density and imaging quality when using a high reconstruction rate. As seen in Fig.~\ref{fig:gray}(c), this results in a temporally variant stream where high-quality frames are interleaved with sparse, low-quality frames, degrading reconstruction stability.

Therefore, it's necessary to automatically retain only the high-quality segments of the event stream. To achieve this, we leverage the real-time acceleration data from the event camera's internal IMU as a prior. Because the acceleration is rigidly coupled to the vibration mechanism, it could provide a precise estimation of the elastomer's current vibration phase and, by extension, the expected imaging quality in real-time. We use this information to develop a model-based algorithm that guides a temporal gate: the filter actively retains the event stream during high-quality phases and discards the stream during low-quality phases. This process ensures the final reconstructed images are built only from high-quality data with enhanced consistency.

\subsubsection{\textbf{Mathematical Modeling}}
To establish a model for event-based tactile imaging under active vibration, we start by analyzing the relationship between vibrational contact and the quality of reconstructed images.
For simplification, we analyze the vibration pattern using only vertical vibration. The vibration excitation model of the elastomer can be expressed as a sinusoidal function. During tactile interactions between the elastomer and external objects, several factors including material properties, deformation characteristics, and system hysteresis would cause variations in amplitude and introduce a phase delay.  The resulting vibration model can be modelled as:
\begin{equation}
f(t)=A\sin(\omega t+b),
\end{equation}
where \( f \), \( A \), \( \omega \), \( t \), and \( b \) are the displacement, amplitude, the angular frequency, time, and the system's phase delay, respectively.

In the imaging model of event cameras, we use an abstract metric, imaging quality (IQ), to analyze the performance. Specific metrics for evaluation will be provided in the next section. The characteristic of event cameras is to capture changes in illumination, and therefore, IQ tends to increase with the absolute vibration velocity of the imaging medium, i.e., the elastomer. Therefore, we assume a proportional relationship for simplicity in modeling as:
\begin{equation}
\label{eq:iq}
\text{IQ}(t)\approx u\left|\frac{\text{d}f(t)}{\text{d}t}\right|=|uA\omega\cos(\omega t+b)|.
\end{equation}
where \( u \) is a scaling factor. By trigonometric identities and Fourier series, the above equation can be rewritten in a standard sinusoidal form:
\begin{equation}
\text{IQ}(t)=\frac{2|uA\omega|}{\pi}+\frac{4|uA\omega|}{3\pi}\sin(2\omega t+2b+\frac{\pi}{2}),
\end{equation}
which indicates that the imaging quality signal contains a sinusoidal component, a phase delay, and a constant offset.

In practice, the velocity is difficult to measure directly, and acceleration data from the IMU are typically used instead. The IMU measurement corresponds to the second derivative of displacement:
\begin{equation}
\text{IMU}(t)=\frac{\text{d}^2f(t)}{\text{d}t^2}+\epsilon(t) =A\omega^2\sin(\omega t+\pi)+\epsilon(t).
\end{equation}
where the $\epsilon(t)$ term, representing the noise and non-periodic elements, which could be effectively handled by band-pass filtering.

At this point, we can make an inference between the vibration and imaging models. Comparing the IQ and IMU signals, two important relationships are utilized:

\begin{itemize}
    \item The signal frequency of IQ is explicitly twice that of the IMU signal.
    \item There is a stable but implicitly phase difference between the two signals. 
\end{itemize}

%This phase difference is influenced by factors such as material hysteresis and data transmission latency. 
Therefore, a temporal alignment between the two signals is required to estimate the optimal time shift $\hat{\Delta t}$. The theoretical relationship between the filtered IMU signal and the imaging quality can thus be approximated as:
\begin{equation}
\text{IQ}(t) = k\cdot\left|\tilde{\text{IMU}}(t-\hat{\Delta t})\right| + c,
\label{regg}
\end{equation}
where $k$ is a scaling factor and $c$ is a constant offset, determined experimentally. Here, $\tilde{\text{IMU}}(\cdot)$ represents the IMU signal component obtained after applying a band-pass filter centered at the active vibration frequency $\omega$.

\begin{figure*}[!t]
\centering
\includegraphics[width=\linewidth]{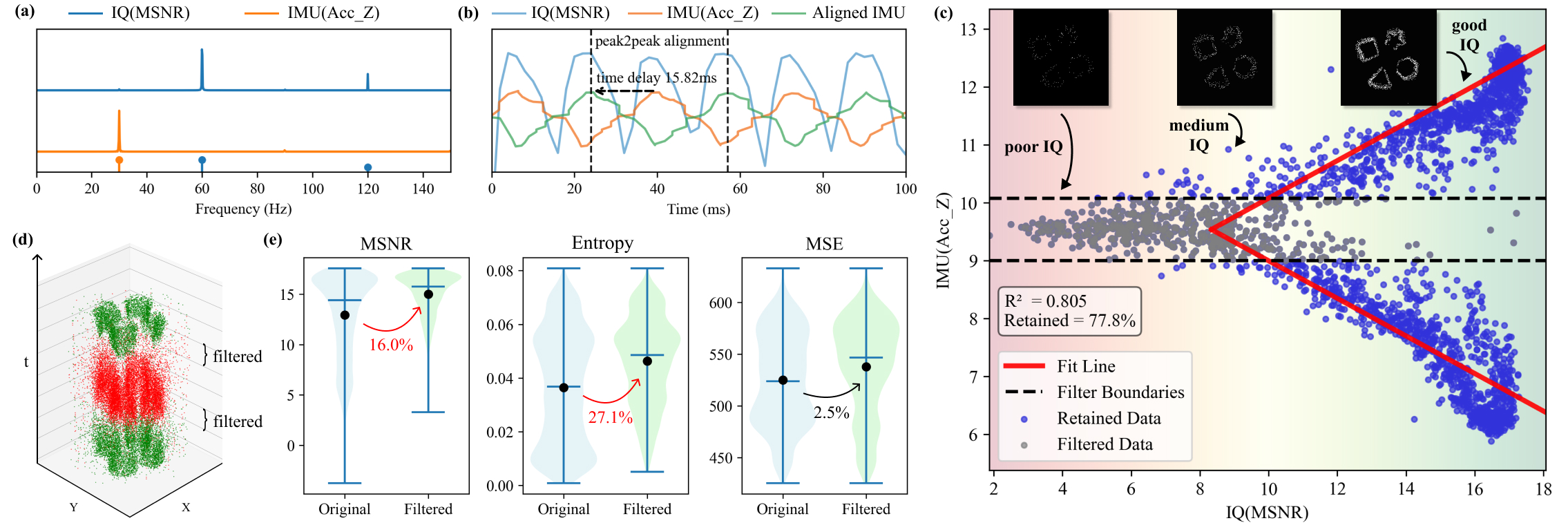}%
\caption{Validation of the IMU‑guided temporal filter. (a) Fourier transform analysis of the imaging quality metric alongside IMU signal.
(b) Peak‑to‑peak alignment.
(c) Relations between IMU and MSNR, illustrating the regression result, filter boundaries and representing frames.
(d) Intuitive illustration of IMU‑guided temporal filtering on the event stream.
(e) Statistical comparison of imaging quality metrics before and after filtering.}
\label{fig_imu}
\end{figure*}

\subsubsection{\textbf{Verification}}
Based on the theoretical analysis, we performed frequency validation, phase alignment, regression fitting to implement the calibration process of the IMU-guided temporal filter. We used data acquired under a 30 Hz vertical vibration with a reconstruction rate of 500 Hz. MSNR is used as the representing imaging quality metric and computed for each image, forming a time series IQ$(t)$ of the total 2500 grayscale images in 5 s. In parallel, the IMU data, along with their corresponding timestamps, were extracted from the raw event camera output. We use the Z‑axis of the IMU data, which represents the vertical vibration, denoted as Acc\_Z. 

First, we applied Fourier transform analysis to both the MSNR series and the IMU measurements. Fig.~\ref{fig_imu}(a) shows that the dominant frequency in the imaging quality metric is 60 Hz, precisely twice of the 30 Hz vibration frequency in the IMU signal, which conforms to our theoretical predictions. Next, we aligned the timestamps of the MSNR series and the IMU signal by a peak-to-peak matching method. We first detecting the peak locations of both IMU and imaging quality signals, and then pairing each IMU peak with its nearest imaging quality peak to determine the average temporal delay \(\hat{\Delta t}\):
\begin{equation}
\hat{\Delta t} = \frac{1}{N}\sum_{n=1}^{N}\left(t_{\text{IQ peak matched}}^{(n)} - t_{\text{IMU peak}}^{(n)}\right),
\label{eq:align}
\end{equation}
As depicted in Fig.~\ref{fig_imu}(b), the 100 ms window shows the frequency doubling and time shift characteristics, verifying the temporal relationship. 

Once the signals are temporally aligned, we pair each reconstructed frame’s MSNR metric with the temporally matched IMU measurement. The images with higher MSNR visually have more complete and distinct edges. Specifically, we calculate the mean of the IMU measurements in the Z-axis over the corresponding interval of accumulation. As shown in Fig.~\ref{fig_imu}(c), the scatter plot of MSNR versus the corresponding IMU measurements displays a clear correlation. A regression analysis was performed based on Eqn.~\eqref{regg}, yielding an $R^2$ value of 0.805. This result validates our theoretical analysis and demonstrates that the IMU measurements can reliably predict the imaging quality of reconstructed frames.

\subsubsection{\textbf{Deployment Strategy}}

For the purpose of filtering, a minimum acceptable imaging quality threshold ($\text{IQ}_{\text{thresh}}$) is defined. In this experiment, for instance, the average MSNR measurement of 10.00 is adopted. Based on the regression analysis of the calibration data, this $\text{IQ}_{\text{thresh}}$ is mapped to a corresponding set of IMU measurement thresholds. Specifically, it's more than 10.08 or less than 9.00 in this example. Using this criterion, poor-quality segments of the event stream can be identified and filtered out, as intuitively illustrated in Fig.~\ref{fig_imu}(d). The result in Fig.~\ref{fig_imu}(c) highlights the retained frames in blue, where 77.8\% of the data were preserved.

The procedure of the IMU-guided temporal filter is summarized in Algorithm \ref{alg:imu_filter}. Rather than directly accumulating events, the IMU measurements embedded in the event camera were used as guidance to filter out the poor streams. It operates in two distinct phases: a one-time calibration and a real-time application. For calibration, it determines the optimal temporal alignment $\hat{\Delta t}$ (Eqn.~\eqref{eq:align}) and the IQ (Eqn.~\eqref{eq:MSNR}) threshold $\text{IQ}_{\text{thresh}}$ based on the reconstructed frames (Eqn.~\eqref{eq:Gk}). Through regression analysis (Eqn.~\eqref{regg}), this quality threshold is then mapped to a corresponding motion magnitude threshold, $\tilde{\text{IMU}}_{\text{thresh}}$. Following calibration, the main filter iterates through the remaining event stream. It interpolates the filtered and aligned IMU value ($\tilde{\text{IMU}}_{\text{live}}$) and compares its magnitude to $\tilde{\text{IMU}}_{\text{thresh}}$. Only segments that satisfy the criterion ($|\tilde{\text{IMU}}_{\text{live}}| \ge \tilde{\text{IMU}}_{\text{thresh}}$) are reconstructed (Eqn.~\eqref{eq:Gk}) and outputted, effectively discarding segments predicted to have poor imaging quality.

\begin{algorithm}[htp]
\caption{IMU-Guided Temporal Filter.}\label{alg:imu_filter}

\begin{algorithmic}
\STATE \textsc{Calibrate}($\mathcal{E}_{\text{cal}}$, $\text{IMU}_{\text{cal}}$, $f_{\text{rate}}$, $\omega$)
    \STATE \hspace{\algorithmicindent} $({G}_{\text{cal}}, {T}_{\text{frames}}) \gets$ Reconstruction($\mathcal{E}_{\text{cal}}$, $f_{\text{rate}}$) by Eqn. \eqref{eq:Gk}
    \STATE \hspace{\algorithmicindent} $\text{IQ}_{\text{series}} \gets$ Calculate\_IQ(${G}_{\text{cal}}$) by Eqn. \eqref{eq:MSNR}
    \STATE \hspace{\algorithmicindent} $\text{IQ}_{\text{thresh}} \gets \text{Mean}(\text{IQ}_{\text{series}})$
    \STATE \hspace{\algorithmicindent} $\tilde{\text{IMU}}_{\text{signal}} \gets$ Bandpass($\text{IMU}_{\text{cal}}$, $\omega$)
    \STATE \hspace{\algorithmicindent} $\tilde{\text{IMU}}_{\text{series}} \gets$ Sample($\tilde{\text{IMU}}_{\text{signal}}$, ${T}_{\text{frames}}$)
    \STATE \hspace{\algorithmicindent} $\hat{\Delta t} \gets$ Alignment($\text{IQ}_{\text{series}}$, $\tilde{\text{IMU}}_{\text{series}}$, ${T}_{\text{frames}}$) by Eqn. \eqref{eq:align}
    \STATE \hspace{\algorithmicindent} $(k, c) \gets$ Regression($\text{IQ}_{\text{series}}$, $\tilde{\text{IMU}}_{\text{series}}$, $\hat{\Delta t}$) by Eqn. \eqref{regg}
    \STATE \hspace{\algorithmicindent} $\tilde{\text{IMU}}_{\text{thresh}} \gets (\text{IQ}_{\text{thresh}} - c) / k$
    \STATE \hspace{\algorithmicindent} \textbf{return} $\hat{\Delta t}$, $\tilde{\text{IMU}}_{\text{thresh}}$
\STATE
\STATE {\textsc{ImuGuidedFilter}}($\mathcal{E}_{\text{stream}}$, $\text{IMU}_{\text{stream}}$, $f_{\text{rate}}$, $\omega$)
    \STATE \hspace{\algorithmicindent} $(\mathcal{E}_{\text{cal}}, \text{IMU}_{\text{cal}}) \gets$ GetFirstSegment($\mathcal{E}_{\text{stream}}$, $\text{IMU}_{\text{stream}}$)
    \STATE \hspace{\algorithmicindent} $(\hat{\Delta t}, \tilde{\text{IMU}}_{\text{thresh}}) \gets$ \textsc{Calibrate}($\mathcal{E}_{\text{cal}}$, $\text{IMU}_{\text{cal}}$, $f_{\text{rate}}$, $\omega$)
    \STATE \hspace{\algorithmicindent} \textbf{while} {$\mathcal{E}_{\text{stream}}$ has remaining segments}
        \STATE \hspace{2\algorithmicindent} $t_{\text{live}} \gets$ Get\_Timestamp($\mathcal{E}_{\text{stream}}$)
        \STATE \hspace{2\algorithmicindent} $\tilde{\text{IMU}}_{\text{live}} \gets$ Sample\&Bandpass(${\text{IMU}}_{\text{stream}}$, $t_{\text{live}} - \hat{\Delta t}$)
        \STATE \hspace{2\algorithmicindent} \textbf{if} {$|\tilde{\text{IMU}}_{\text{live}}| \ge \tilde{\text{IMU}}_{\text{thresh}}$}
            \STATE \hspace{3\algorithmicindent} $G_{\text{live}} \gets$ Reconstruction($\mathcal{E}_{\text{stream}}$, $f_{\text{rate}}$) by Eqn. \eqref{eq:Gk}
            \STATE \hspace{3\algorithmicindent} \textbf{Output} $G_{\text{live}}$
        \STATE \hspace{2\algorithmicindent} \textbf{end if}
    \STATE \hspace{\algorithmicindent} \textbf{end while}
\end{algorithmic}
%}
\end{algorithm}

Statistical analysis comparing the pre-filtered and post-filtered results (Fig.~\ref{fig_imu}(e)) shows that the IMU-guided temporal filter substantially improves the MSNR and entropy, while MSE has a slight increment. Specifically for MSNR, an increase in the average value (Avg.) corresponds to an overall enhancement of imaging quality, while a reduction in the standard deviation (Std.) reflects improved stability and temporal consistency. Further experiments across various vibration frequencies and reconstruction rates (Table~\ref{tab:msnr_retention}) demonstrate a maximum average increase of 24.0\% and a standard deviation decrease of 45.8\% in MSNR.

% For example, in our 30 Hz dataset with 1000 Hz reconstruction rate, the filter retained 77.3\% of the frames, corresponding to an effective frame rate of 773 Hz. In contrast, reconstructing the event stream at 773 Hz without this filter would include many low-quality frames. Additionally, the reconstructed frames with IMU-guided temporal filter exhibit a minimal 1 ms delay, underscoring the benefit of this strategy.

\begin{table}[htbp]
  \centering
  \renewcommand{\arraystretch}{1.2} % 增大行间距
  \caption{MSNR enhancement by IMU-guided temporal filter}
  \resizebox{\columnwidth}{!}{
  \begin{tabular}{cccccc}
    \toprule
    \makecell[c]{Vibration\\Frequency}  & \makecell[c]{Reconstru-\\ction Rate} & \makecell[c]{Processing\\Delay}  & \makecell[c]{Retaining\\Rate} & \makecell[c]{Avg.\\Increase} & \makecell[c]{Std.\\Decrease}  \\
    \midrule
    30 Hz    & 500 Hz  & 2 ms  & 77.8\% & 16.0\% & 45.8\% \\
    30 Hz    & 1000 Hz & 1 ms  & 81.1\% & 24.0\% & 40.8\% \\
    50 Hz    & 1000 Hz & 1 ms  & 93.0\% & 11.6\% & 23.0\% \\
    \bottomrule
  \end{tabular}
  }
  \label{tab:msnr_retention}
\end{table}

% Overall, our proposed IMU-guided temporal filtering strategy is summarized as follows. Rather than directly accumulating events, the IMU measurements embedded in the event camera should be read in advance. When the IMU measurement fall within the defined threshold range, the event stream is triggered for accumulation; otherwise, it will be kept for the next accumulation. The IMU‐guided temporal filter retains most reconstructed frames with a minimum processing delay, enhancing the quality and temporal consistency of event-based tactile images. 

\subsection{Contact Surface Estimation}

\begin{figure}[t]
    \centering
    \includegraphics[width=\linewidth]{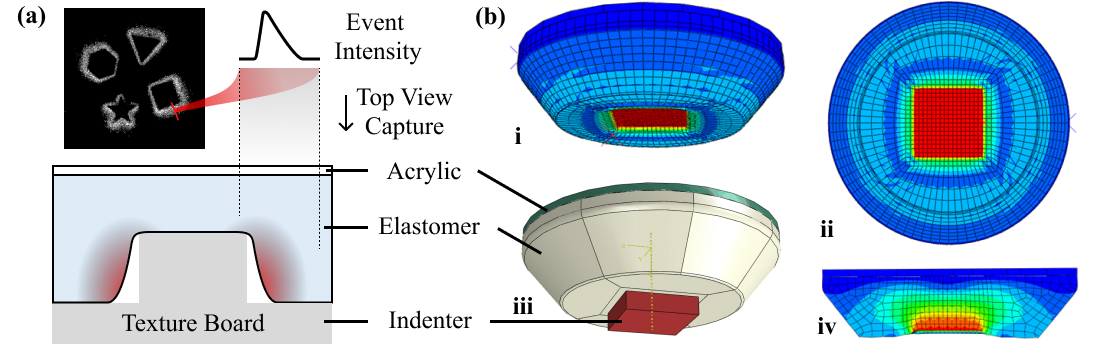}
    \caption{Motivation and method of contact surface estimation. (a) Elastomer deformation under compression captured as asymmetric edge features. (b) Finite element analysis of elastomer deformation. iii. Simulation setup. Deformation contour plot in i. isometric; ii. top; iv. cross-sectional view of the elastomer.}
    \label{fig:contact}
\end{figure}

In Section~\ref{sec:calib}, we noted that the event camera only highlights the edges of contacting objects. As a result, the reconstructed image underestimates the interior contact region and fragments the physical footprint. Therefore, it's necessary to estimate the full contact surface from the tactile image~\cite{tex1,tex2,taunyazov_event-driven_2020}. The capture edges arise from a relative height difference on the texture board. Due to the elastomer’s continuity and elasticity, it cannot conform exactly to the vertical face of the texture, as illustrated in Fig.~\ref{fig:contact}(a). Instead, the deformation is governed by Young’s modulus and Poisson’s ratio~\cite{OLSSON2019985,MOISIO20131}. It radiates outward from the contact edge until the stress relaxes. In the protrusion scenario, the deformation of elastomer creating a shadow that extends beyond the true boundary, while the contact side remains sharp. The event intensity curve clearly illustrates the asymmetric feature, with one side sharpened while the other blurred. 

To further validate this asymmetric deformation, we performed finite element analysis (FEA) simulations using Abaqus. The model depicted in Fig.~\ref{fig:contact}(b)iii featured a rectangular rigid indenter pressing onto the elastomer surface with 2~mm vertical displacement, employing surface-to-surface contact conditions. A fixed rigid  acrylic plate fully constrains the elastomer from above. Elastomer parameters were set to a Young’s modulus of 0.2~MPa and a Poisson’s ratio of 0.48, using 1.5~mm C3D8R elements for meshing. The resulting deformation contour plot in Fig.~\ref{fig:contact}(b) confirms the asymmetrical edge deformation. Stress rapidly accumulates near the inner contact boundary, creating an abrupt deformation edge (sharp imaging), while it gradually dissipates outward, resulting in a smooth deformation gradient (blurred imaging). The close agreement between simulation results and experimental imaging validates our assumption about asymmetric deformation patterns. Therefore, without needing markers, we can leverage this feature to recover the full geometry of the contact surface rather than merely estimating its area~\cite{yin_geleventnovel_2025,s140405805}.

\subsection{Sensor Characterization}

To characterize the SWTac sensor, we performed a statistical analysis on grayscale tactile images reconstructed from event streams. We systematically evaluated SWTac's performance by varying key parameters, including: vibration amplitude (0–400 \textmu m), frequency (0–400 Hz), event sensitivity (three thresholds), vibration direction (vertical and horizontal) and elastomer property (viscoelasticity). A constant reconstruction rate of 100 Hz was used throughout the characterization.

\begin{figure*}[ht]
    \centering
    \includegraphics[width=1\linewidth]{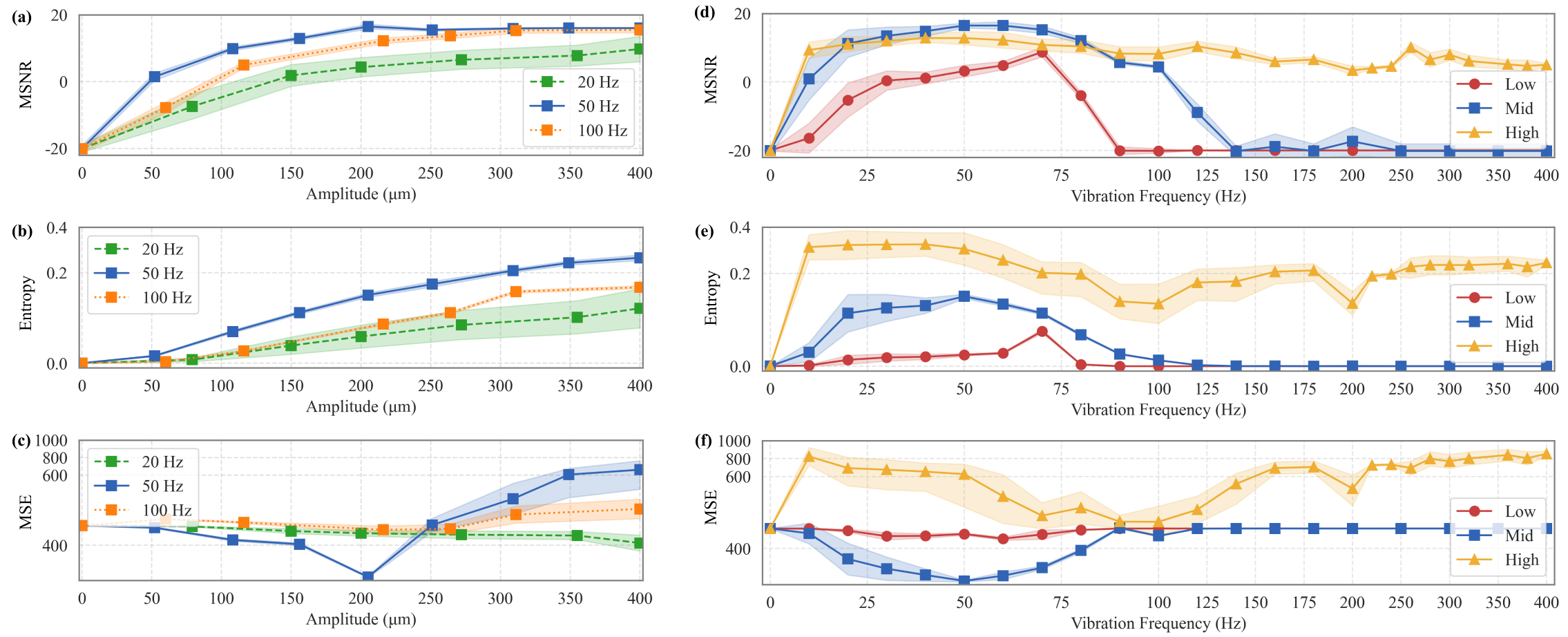}
    \caption{Statistical evaluation of grayscale imaging quality under different vibrational conditions.
(a–c) Amplitude variations with a mid sensitivity setting and three different frequencies.
(d–f) Frequency variations with a 200 \textmu m amplitude and three different event sensitivity settings.}
    \label{fig:freq}
\end{figure*}

\subsubsection{\textbf{Vibration Amplitude}}

First, we evaluated the sensor's performance variation with amplitude at three different frequencies, as shown in Fig.~\ref{fig:freq}(a-c). The results indicate that when the amplitude is below 100 \textmu m, the MSNR values are low but increase rapidly. Above 200 \textmu m, the MSNR remains stable, but its variance increases. Notably, a medium vibration frequency of 50 Hz consistently provides superior performance.

The rationale for this behaviour is twofold. For lower amplitudes, the induced deformation is too small to generate sufficient information. While higher amplitudes initially capture more information (increasing entropy), amplitudes up to 200 \textmu m yield signals that more closely approximate the ground truth (indicated by reduced MSE). Further increases in amplitude cause the signal to deviate due to higher reaction forces and noise. This is because the larger amplitude exceeds the elastomer's effective deformation range. Consequently, the apparent Young's modulus increases sharply, leading to smaller strain changes despite larger deformations. In short, although larger amplitudes capture more information, the added noise simultaneously reduces imaging quality.

\subsubsection{\textbf{Vibration Frequency}}
\label{sec:freq}

Next, we evaluated the sensor's performance under various vibration frequencies (0–400 Hz), as shown in Fig.~\ref{fig:freq}(d-f). For frequencies between 0–100 Hz, the amplitude was kept at a constant 200 \textmu m. While for frequencies between 100 and 400 Hz, the power amplification factor was fixed due to constraints of the IMU range. In the low-frequency range (0–100 Hz), the MSNR first gradually improves with increasing frequency, peaking at about 50 Hz. After this peak, the MSNR declines. At higher frequencies (100–400 Hz), damping effects become significant. For mid and low sensitivity, entropy drops to nearly zero, indicating no effective information is captured, and MSE approaches levels seen without vibration (0 Hz). For high sensitivity, MSNR remains relatively stable but is still worse than in the low-frequency range.

The initial performance increase up to 50 Hz is because the masked ground truth region accumulates more events, resulting in clearer edge signals (evidenced by higher entropy and lower MSE). The subsequent decline occurs because the elastomer's damping effect begins to dominate over the active vibration. At higher frequencies, these damping effects and limited elastomer deformation cannot be fully compensated by post-processing, making high-frequency vibration unsuitable. Specifically, in the 20–40 Hz range, the higher standard deviation compared to 50 Hz is observable. This is attributed to the variance in reconstructed image, which arises from capturing the elastomer in different vibration phases. This is precisely the issue addressed in Section~\ref{sec:imu}.

\subsubsection{\textbf{Event Camera Sensitivity}} 

The selection of the event threshold presents a clear trade-off in Fig.~\ref{fig:freq}(d-f). Our results show that the highest MSNR is achieved at approximately 50 Hz with a mid sensitivity setting. In the 20–80 Hz frequency range, the mid sensitivity consistently outperforms high sensitivity. At higher frequencies, although a higher sensitivity can capture texture patterns, the imaging quality still does not match the performance observed in the low-frequency range.

This trade-off is explained by the other metrics: a higher event threshold captures more information, reflected by higher entropy. However, it also produces thicker, more pronounced edges and increased background noise, leading to a higher MSE. In contrast, lower sensitivity reduces noise but may result in incomplete edge detection. The mid sensitivity setting, therefore, provides the optimal balance between information capture and noise induction.

\subsubsection{\textbf{Vibration Direction}}

\begin{figure}[t]
    \centering
    \includegraphics[width=\linewidth]{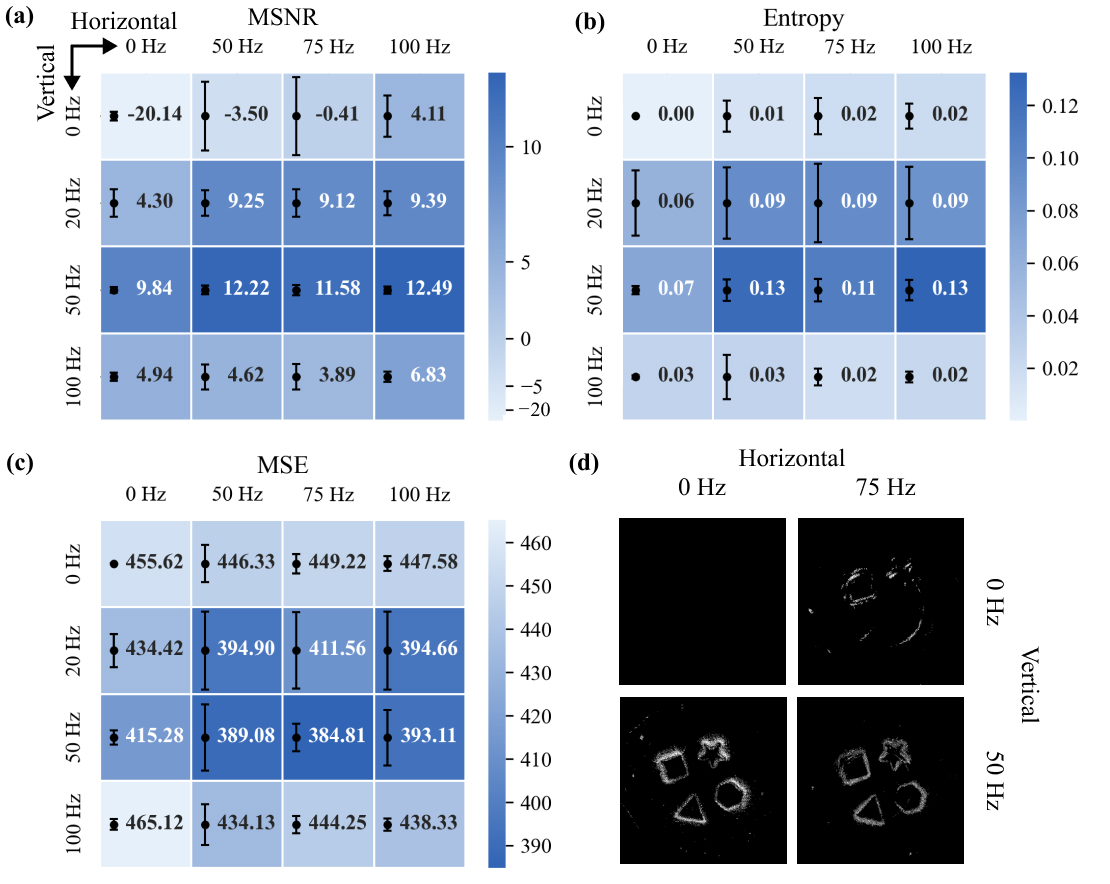}
    \caption{Grayscale imaging quality under different vibration directions.
(a–c) Imaging quality metrics under different frequencies.
(d) Illustration of reconstructed grayscale images.}
    \label{fig:mode}
\end{figure}

We evaluated the effect of superimposing horizontal vibration (0, 50, 75, 100 Hz) onto vertical vibration (0, 20, 50, 100 Hz). The results in Fig.~\ref{fig:mode}(a-c) show that introducing horizontal vibration generally enhances the MSNR, increases entropy, and reduces MSE. This improvement is especially significant when vertical vibration is less effective (e.g., at 0 Hz). An intuitive illustration is provided in Fig.~\ref{fig:mode}(d).

The improvement is attributed to the fact that horizontal vibration leads to more uniform contact, enhancing edge integrity in the reconstruction images. Although the amplitude of horizontal vibration is small, it effectively compensates when vertical vibration performance is suboptimal. Moreover, horizontal vibration helps release stress accumulated from vertical oscillation, contributing to structural stability mechanically. In summary, the combined 100 Hz horizontal and 50 Hz vertical vibrations yield the best overall performance.

\subsubsection{\textbf{Elastomer Viscoelasticity}}
\label{elas}

The performance peak observed in our experiments is closely related to the material properties of the PDMS elastomer. Theoretically, PDMS is a viscoelastic material whose mechanical behavior are primarily influenced by cross-linking ratio~\cite{teixeira_polydimethylsiloxane_2021} and material thickness~\cite{trindade_modeling_2000}. It exhibits time-delayed responses and energy dissipation, and in vibrational environments, the strain-rate effect further reduces the effective elastic range at higher frequencies~\cite{kumar}. Although the transient response of elastomers is studied in acoustics (typically $>$1000 Hz)~\cite{trindade_modeling_2000} and biomedical applications (around 1 Hz)~\cite{kim_development_2021}, experimental data in the mechanical vibration range (10–100 Hz) remains scarce. Therefore, we adopted a modeling approach to illustrate the general trend. We used the Kelvin-Voigt model to describe this behavior, whose constitutive equation is:
\begin{equation}
\sigma(t) = E\varepsilon(t) + \eta\frac{\text{d}\varepsilon(t)}{\text{d}t}.
\end{equation}

\begin{figure}[t]
    \centering
    \includegraphics[width=\linewidth]{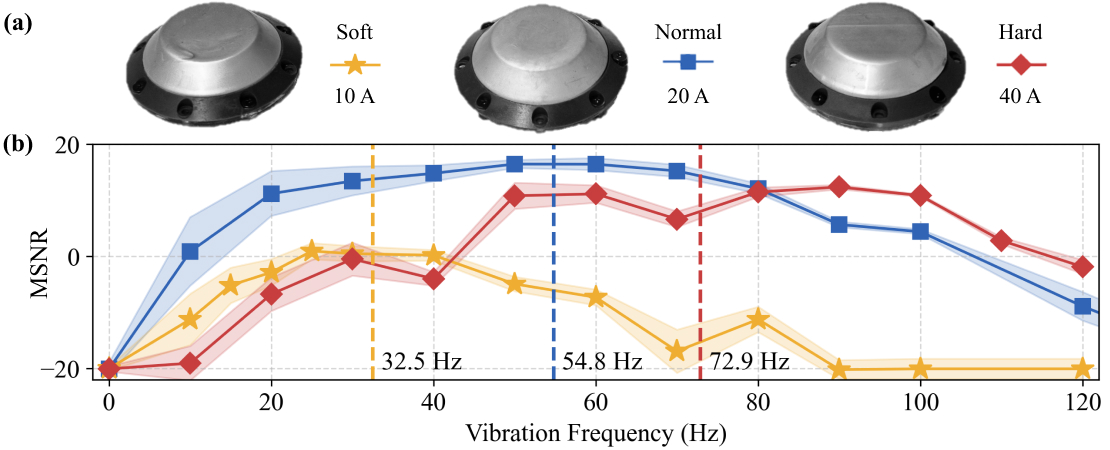}
    \caption{Grayscale imaging quality for elastomers of different {Shore hardness, representing different viscoelasticity}. (a) Photographs of elastomers. (b) Imaging quality (e.g. MSNR) versus vibration frequency.}
    \label{fig:soft}
\end{figure}

This model shows the elastomer acts as a first-order low-pass filter. At lower frequencies ($\omega \to 0$), it exhibits a near-ideal elastic response (gain $\to 1/E$), but at higher frequencies ($\omega \to \infty$), viscous damping dominates, causing significant amplitude reduction and phase lag. The characterization result in Fig.~\ref{fig:freq}(d), showing a clear MSNR decline in the 50–120 Hz range, aligns with this model.

To confirm this, we fabricated and tested three elastomers with different hardness levels (Fig.~\ref{fig:soft}(a)). The results in Fig.~\ref{fig:soft}(b) show that as elastomer hardness increases, its elastic response is enhanced, and the MSNR peak frequency also rises. This confirms the relationship between material viscoelasticity and the damping behavior transition. The elastomer with a hardness of approximately 20 A (17:1 PDMS formulation) consistently delivered superior performance, balancing signal inconsistency at lower frequencies and energy loss at higher frequencies. The softer elastomer failed to capture texture details, while the harder one required higher contact forces, degrading performance.

\section{Experiments}
\label{sec:exp}
In this section, we first validate vibration designs in SWTac and the sensor performance on core tactile tasks: contact surface estimation, force estimation, and stone classification. We then evaluate the integrated SandWorm robot, assessing its penetration, autonomous locomotion, pipeline dredging, and subsurface exploration capabilities. Finally, we demonstrate the system's real-world applicability through locomotion and perception field tests.

\subsection{Vibration System Validation}
\label{sec:exphard}
\begin{figure}[t]
    \centering
    \includegraphics[width=\linewidth]{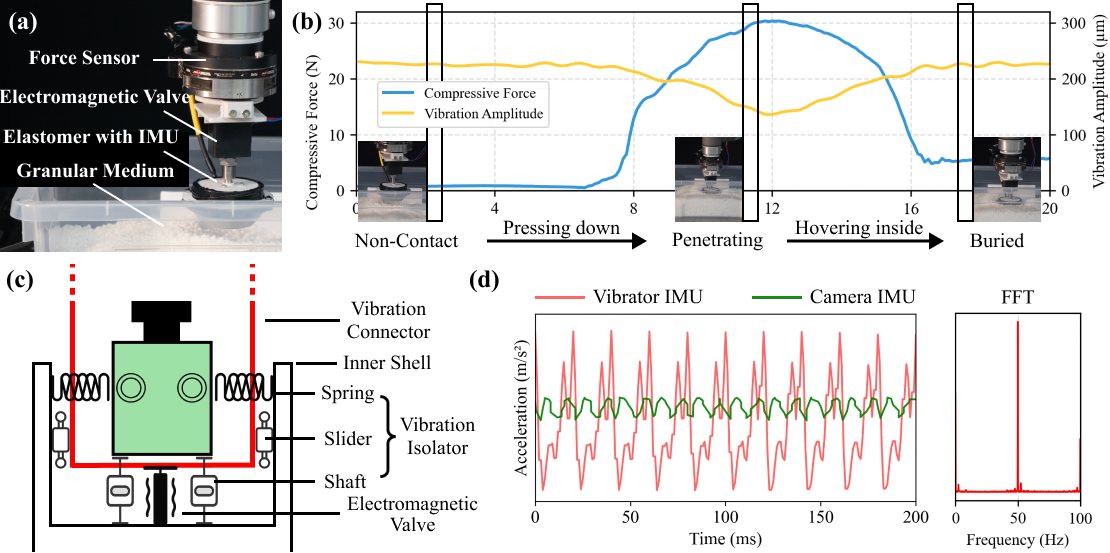}
    \caption{Hardware validation of the onboard vibration system. (a) Test setup to measure vibration amplitude versus compressive force. (b) Evolution of compressive force and vibration amplitude during penetration into granular media. (c) Structural schematic of the vibration isolation mechanism. (d) IMU measurements for isolation and FFT analysis.}
    \label{fig:vib}
\end{figure}

In this experiment, we validated the vibration design of SWTac. First, we tested the performance of the onboard electromagnetic valve to ensure that its active vibration remains effective under actual compressive loads. As shown in Fig.~\ref{fig:vib}(a), we constructed a test platform with a force sensor to measure compressive load and an external IMU attached to the elastomer to measure vibration. We examined the amplitude-frequency characteristics under different contact force conditions. The results indicate that the vibration frequency consistently remained stable at 50 Hz, regardless of the contact force. Fig.~\ref{fig:vib}(b) shows that as the elastomer is pressed and hovered within the granular medium, the vibration amplitude decreases with increasing contact pressure. However, even under a high compressive load of 30 N, the electromagnetic valve maintained an amplitude of 134 \textmu m. While this reduces the MSNR by approximately 20\% from its maximum, the resulting signal remains sufficiently clear for effective perception. We further confirmed that under other normal pressure conditions, such as during rotation and in different granular media, the amplitude also remains well within the ideal operating range.

Furthermore, we verified the effectiveness of the hardware-level vibration-isolation mechanism, which is crucial for stable imaging. As shown in Fig.~\ref{fig:vib}(c), the electromagnetic valve actuates the elastomer through the vibration connector, while the event camera is mounted to the inner shell via several vibration isolators. To validate this structure, we modified the SWTac to attach an IMU to the vibration connector, and thereby compare its data to the event camera's internal IMU. The results in Fig.~\ref{fig:vib}(d) confirm an 83\% reduction in acceleration transmitted to the camera, proving the hardware decoupling significantly attenuates vibration. Additionally, the FFT analysis shows the vibration frequency remains precisely locked at 50 Hz with an error of less than 0.2\%. This stability originates from the MOSFET switch controlling the electromagnetic valve.

To this end, based on the characterization results and this hardware validation, we established the sensor configuration used for all subsequent experiments: a 17:1 PDMS elastomer, a 50 Hz vertical vibration from the electromagnetic valve, a 100 Hz horizontal vibration, and a mid-level event sensitivity setting.

\begin{figure*}[t]
    \centering
    \includegraphics[width=0.98\linewidth]{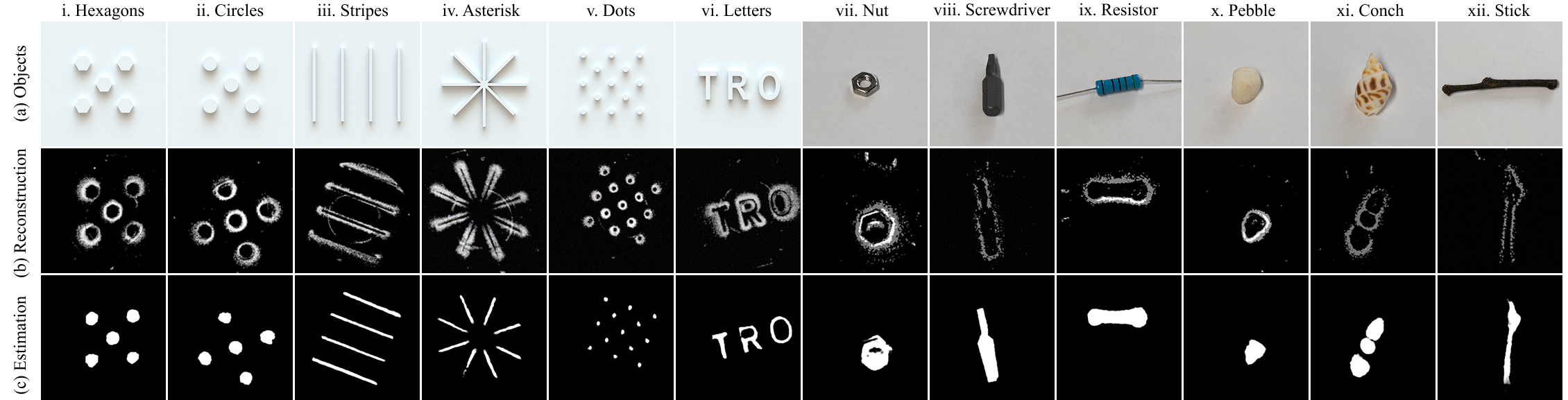}
    \caption{Contact surface estimation. (a) Texture boards and real-world objects. (b) Raw accumulated grayscale images. (c) Estimation result by U-Net.}
    \label{fig:net}
\end{figure*}

\subsection{Contact Surface Estimation}
Motivated by these intuitive asymmetric edge features, we adopted a U-Net architecture~\cite{unet} to automatically recover the contact surface from raw accumulated edge images. This architecture integrates multi-scale context while preserving boundary detail through skip connections. It is well suited to delineate sharp contact boundaries and to suppress blur introduced by lateral elastomer deformation. We adopt a 4-level U-Net, which first maps a single-channel grayscale input to a single-channel mask. Each encoder stage uses two 3$\times$3 convolutions (BN+ReLU), followed by 2$\times$2 max pooling; the bottleneck repeats the double 3$\times$3 convolutions. The decoder upsamples with 2$\times$2 transposed convolutions and skip connections, each followed by two 3$\times$3 convolutions (batch normalization and ReLU). A final 1$\times$1 convolution with a sigmoid produces the mask.

For the dataset, we used 12 objects shown in Fig.~\ref{fig:net}(a), including six 3D-printed texture boards and six real-world objects.
A total of 300 reconstructed tactile images are collected with manual annotations of the contact region. The accumulated images are shown in Fig.~\ref{fig:net}(b). Moreover, Data augmentation expanded the set to 3000 images via stochastic transforms. For cross-category evaluation, we adopted category-holdout splits: six categories for training and all for testing, with 1000 training and 1000 testing images per split.

For the training process, input images were resized to $256\times256$. The model is trained with an $\ell_1$ loss and Adam optimizer~\cite{adam} (learning rate $1\times10^{-4}$), using a batch size of 64 for 200 epochs. The training process ran on an NVIDIA 5090 GPU with an AMD R9-9950X CPU. Three metrics were used to evaluate the estimation result: structural similarity index measure (SSIM), intersection over union (IoU), and RMSE. SSIM measures the perceptual similarity between reconstruction and ground truth, IoU quantifies the overlap between predicted and true contact regions, and RMSE indicates the average pixel-wise error between the estimation and ground truth. We trained on five randomly selected category-holdout splits and compared models with and without the IMU-guided temporal filter; results are reported in Table~\ref{tab:unet}.

\begin{table}[h]
\centering
\caption{Quantitative Result of Contact Surface Estimation}
{
\begin{tabular}{ccccc}
\toprule
~ & SSIM $\uparrow$ & IoU $\uparrow$ & RMSE $\downarrow$ \\
\midrule
Filtered   & \textbf{0.9688 $\pm$ 0.0004} & \textbf{0.8104 $\pm$ 0.0052}  & \textbf{0.0693 $\pm$ 0.0011} \\
Raw    & 0.9684 $\pm$ 0.0043 & 0.6965 $\pm$ 0.0077  & 0.0945 $\pm$ 0.0016 \\
\bottomrule
\end{tabular}
\label{tab:unet}}
\end{table}

Across all splits, U-Net delivers strong estimation performance and IMU-guided filtering consistently outperforms raw inputs. The increase in IoU shows cleaner region delineation and fewer boundary errors. The decrease in RMSE reflects smaller pixel-wise deviations. SSIM remains comparable, which suggests the U-Net already captures global structure. These gains indicate that filtering suppresses vibration-induced jitters and stabilizes event density, enabling the U-Net to sharpen boundaries and reject shadowing from lateral elastomer deformation. Fig.~\ref{fig:net}(c) intuitively shows that the predicted masks align closely with the true contact regions, capturing fine contours and small gaps while minimizing spurious halos. For the limitation, the high textural density near the asterisk pattern center leads to insufficient elastomer deformation, making it difficult to accurately reconstruct the modelled geometry. Together, these results validate the proposed algorithm and demonstrate that SWTac provides reliable visuotactile output for marker-free contact geometry recovery.

\subsection{Force Estimation}

\begin{figure}[t]
    \centering
    \includegraphics[width=\linewidth]{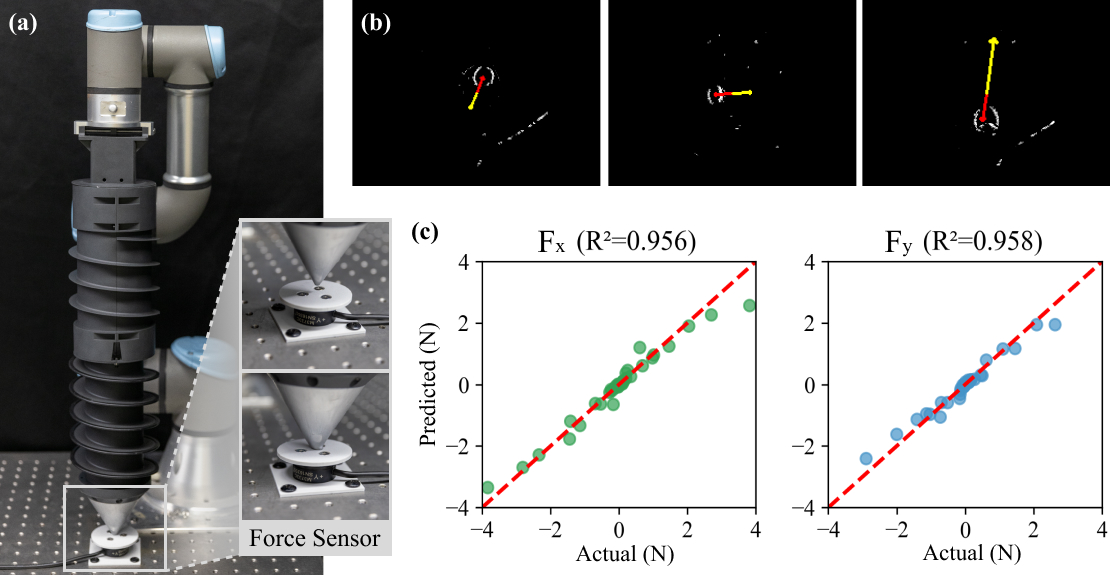}
    \caption{Shear force estimation. (a) Calibration platform highlighting the contact between the elastomer and the force sensor. (b) Illustration of force estimation results.  
(c) Regression results of predicted and actual forces.}
    \label{fig:force}
\end{figure}

In this experiment, we employ a conical elastomer to validate SWTac’s capability to provide force information for downstream tasks. The force can be estimated by tracking the displacement at the tip~\cite{Yuan2017GelSight,funk_evetac_2024,mukashev_e-bts_2025}. We focus on shear measurements because the sensor’s active vibration is primarily in the normal direction, making the normal force strongly oscillatory and of limited practical value. Though it can be estimated robustly via short-time averaging, this comes at the cost of a reduced effective sampling rate. We developed the calibration platform shown in Fig.~\ref{fig:force}(a). By precisely controlling the sensor's lateral offset with a robotic arm, we recorded event streams corresponding to different shear forces. As the offset increases, the tip eventually slides on the force sensor due to insufficient friction, defining the onset of the maximum shear force. A total of 36 data segments were collected for this analysis.

Due to the simplicity of the target geometry, we track the featured tip from the denoised grayscale images using Canny edge detection~\cite{canny} with tuned spatiotemporal constraints. Using the undeformed elastomer tip as the baseline reference, we compute the minimum enclosing circle of the tracked tip and record its central geometry $(x,y)$ relative to this reference. With the collected dataset, we trained regression models to estimate the force vector $(F_x, F_y)$ from the extracted geometric features. An auxiliary feature $r$ is calculated as the Euclidean distance from the feature center to the reference center. For shear forces $F_x$ and $F_y$, we adopted input features $[x, y, r]$ and used the Random Forest~\cite{Breiman2001} regressor for its ability to capture nonlinear dependencies and its robustness on small, noisy datasets. Evaluation on the dataset showed high accuracy, as illustrated in Fig.~\ref{fig:force}(c), with $R^2$ scores exceeding 0.95 for shear forces and mean average error (MAE) below 0.15 N. For each frame, the predicted force vector $(F_x, F_y)$ was overlaid on the image as shown in Fig.~\ref{fig:force}(b). The red arrow represents the displacement of the featured tip from the reference position, indicating direction, while the yellow arrow shows the shear force direction and magnitude. For display purposes, the force direction was inverted.

\subsection{Stone Classification}
\begin{figure}[t]
    \centering
    \includegraphics[width=\linewidth]{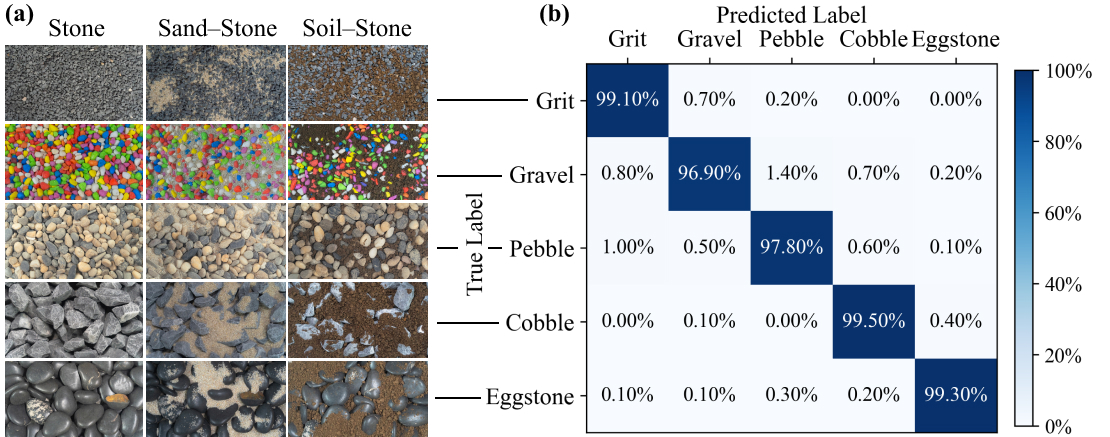}
    \caption{Stone classification. (a) Sample images of different types of stones. (b) Confusion matrix of classification results.}
    \label{fig:rock}
\end{figure}

Accurate classification of stone types, which exhibit unique granularity and surface textures, is essential for subsurface exploration. This experiment tests SWTac's ability of texture and validates its robustness in realistic downstream applications. We collected the tactile data for five stone types in the real world: grit, gravel, pebble, cobble, and eggstone. Data was gathered under three conditions: stone only, sand–stone composites, and soil–stone composites, as shown in Fig.~\ref{fig:rock}(a). The IMU-guided filter was applied to the signals after removing non-contact intervals with reconstruction rate at 500 Hz. Data augmentation was applied to the dataset, yielding a total of approximately 10,000 images.

For classification, we adopted a ResNet-18 model pre-trained on ImageNet~\cite{7780459}. We employed a transfer learning strategy, freezing the stem modules to preserve low-level features. These modules contain the initial convolution, batch normalization, and max pooling. The remaining residual stages were fine-tuned. The final fully connected layer was replaced with a dropout layer and a new linear layer with five outputs. The dataset was split into training and test sets at a 1:1 ratio. The model was trained for 50 epochs using an Adam optimizer with a learning rate of $10^{-4}$, an $\ell_2$ decay of $10^{-4}$, and a batch size of 32 with cross-entropy loss. Training was conducted on an NVIDIA 5090 GPU with an AMD R9-9950X CPU.

The resulting confusion matrix is shown in Fig.~\ref{fig:rock}(b). The model achieved an overall accuracy of 98\%, successfully classifying the stone types even in these unstructured, composite environments. This high robustness is attributed to the sensor's active vibration mechanism. The high-frequency vibrations help dislodge fine particles like sand and soil from the stone's surface, reducing interference and allowing the sensor to capture the underlying tactile signature.

\subsection{Penetration Benchmark}

\begin{figure}[t]
    \centering
    \includegraphics[width=\linewidth]{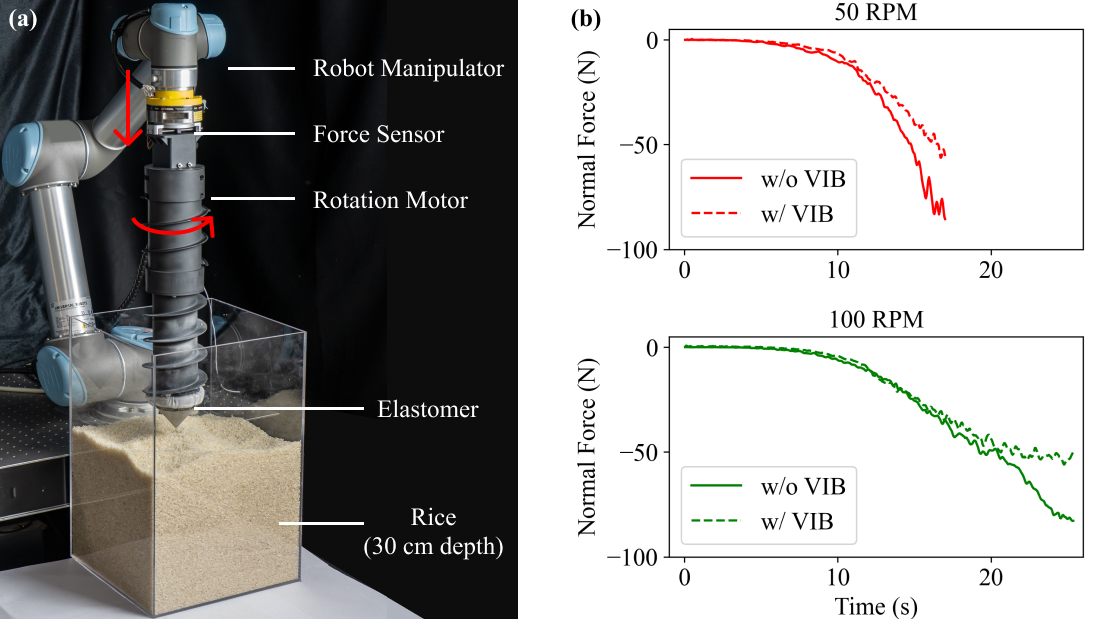}
    \caption{Benchmark of vibration-assisted penetration. (a) Experimental setup. (b) Force sensor readings under static and vibrational excitation at different rotational speeds during penetration.}
    \label{fig:pene}
\end{figure}

In this experiment, we quantify the contributions of the screw thread and active vibration to the penetration performance of the SandWorm robot. To create a standardized test, we constructed a test platform, as shown in Fig.~\ref{fig:pene}(a), using rice as the benchmark medium. The rice was piled approximately 300 mm high in a container. SWTac was fixed to a robotic arm via a high-range force sensor, with the brushless motor providing rotational motion. A soft dust-proof mesh was installed around the elastomer to prevent particle ingress, a measure maintained in all subsequent experiments. The robotic arm descended at a constant speed of 5 mm/s. By varying the rotational speed, we compared the effect of vibration on penetration performance. 

The results in Fig.~\ref{fig:pene}(b) show that higher rotational speeds facilitated easier penetration. Crucially, the presence of vibration consistently led to lower penetration resistance at the same depth across all rotational speeds. At 100 RPM, the sensor penetrated beyond 200 mm with relatively stable force readings. Furthermore, the beneficial effect of vibration was most pronounced at lower rotational speeds. These findings confirm that active vibration significantly enhances penetration efficiency in granular media, validating the effectiveness of the sensor's structural design.

\subsection{Pipeline Inspection}

\begin{figure}[t]
    \centering
    \includegraphics[width=\linewidth]{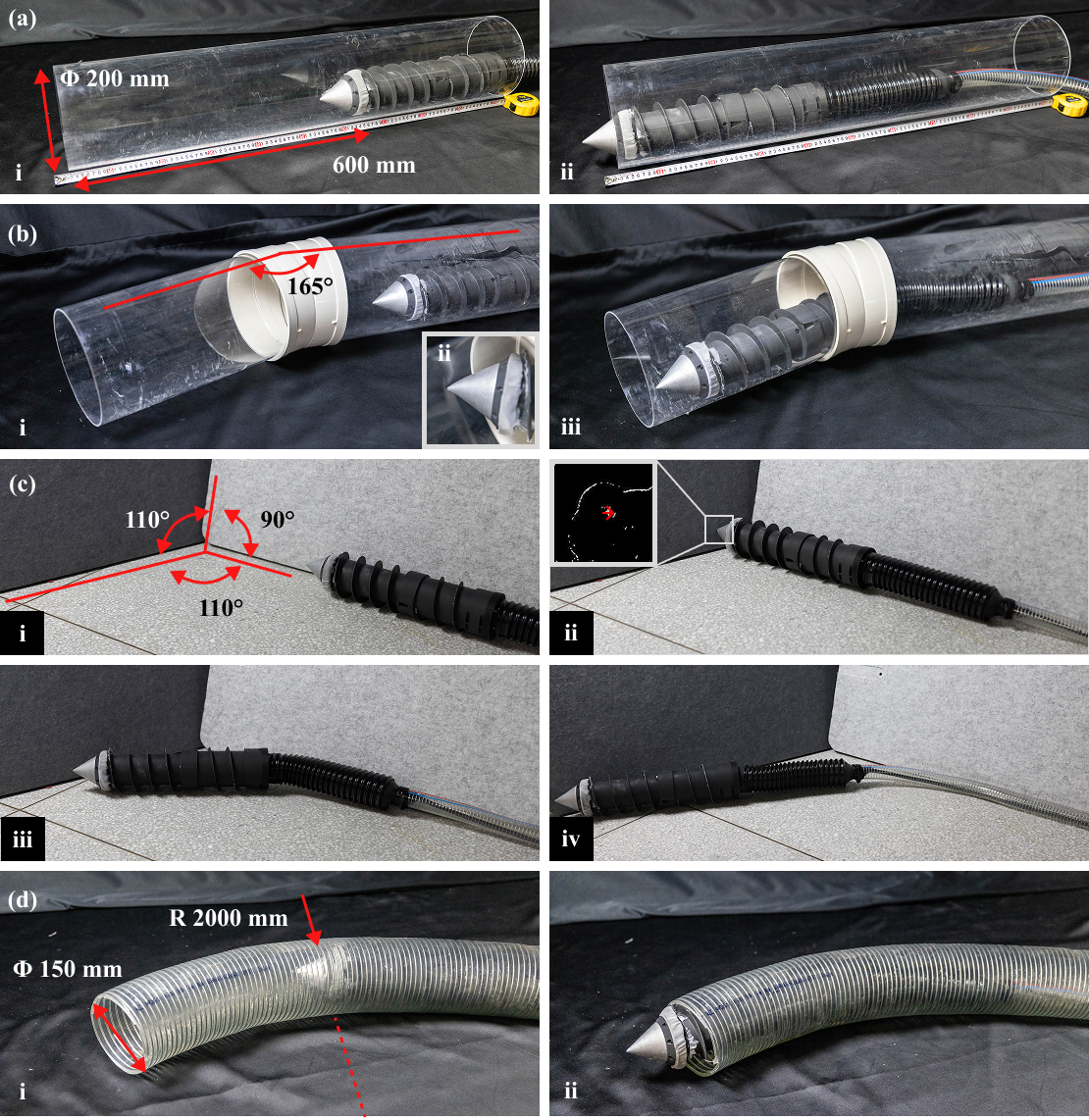}
    \caption{Performance evaluation of the autonomous SandWorm robot in confined environments. (a) Linear locomotion within a straight pipe.  
(b) Steering inside the pipe. 
(c) Steering at wall intersections. 
(d) Traversal through a curved pipe segment.}
    \label{fig:explore}
\end{figure}

% In this task, we first characterized the locomotion performance of SandWorm under four distinct motion modes. As shown in Fig.~\ref{fig:explore}(a), the experiments were conducted in a standardized acrylic pipe fixed to the floor, with an inner diameter of 200 mm and a length of 1 m. The robot’s starting position (Fig.~\ref{fig:explore}(a)i) was set at where the threaded section was fully inside the pipe, and the trial is completed when the elastomer at the tip emerged from the downstream end (Fig.~\ref{fig:explore}(a)ii). The total traversal distance was approximately 600 mm. The pushrod reciprocates over a 30 mm stroke, with each cycle lasting 2 s. We tested two rotational speeds and compared each with and without linear pushrod reciprocation, yielding four motion configurations for evaluation. The results are summarized in Table~\ref{tab:serpentine_performance}.

% The addition of pushrod reciprocation generates a peristaltic effect that reduces traversal time by around 55\% at 60 RPM and 62\% at 90 RPM. This improvement stems from the alternation between extension and retraction phases, overcoming static friction more effectively than rotation only, as analyzed in Section~\ref{sec:kinematics}. SandWorm achieves the maximum straight-line speed of 12.5 mm/s. The synergy between linear reciprocation and screw rotation becomes more pronounced at higher speeds, where inertial coupling further enhances forward propulsion. These results confirm that integrating pushrod motion is critical for efficient locomotion in confined environments.

In this experiment, we first characterized SandWorm's linear locomotion in a 1 m long, 200 mm inner diameter acrylic pipe (Fig.~\ref{fig:explore}(a)). We tested four motion configurations over a 600 mm distance: rotational speeds of 60 and 90 RPM, each with and without the 30 mm pushrod reciprocation. The results are summarized in Table~\ref{tab:serpentine_performance}. It shows that the peristaltic motion by pushrod reciprocation reduces traversal time by up to 62\%. This improvement stems from the alternation of extension and retraction by the pushrod, which overcomes static friction more effectively than screw actuation alone. This synergy yielded a maximum straight-line speed of 12.5 mm/s, confirming that integrated pushrod motion is critical for efficient locomotion.

\begin{table}[ht]
  \centering
  \caption{Traversal time of SandWorm in the pipeline}
  \label{tab:serpentine_performance}
  \begin{tabular*}{0.95\columnwidth}{@{\extracolsep{\fill}} c c c }
    \toprule
    ~~~Configuration                         & 60 RPM & 90 RPM~~~ \\
    \midrule
    ~~~~w/o Pushrod Reciprocation      & 140 s  & 127 s~~~  \\
    ~~~~w/ Pushrod Reciprocation         & 62 s   & 48 s~~~   \\
    \bottomrule
  \end{tabular*}
\end{table}

Next, we evaluated the steering performance of SandWorm in a 15° pipe bend. Fig.~\ref{fig:explore}(b) shows the starting and finishing configurations. For propulsion, we employed a rotational speed of 90 RPM combined with pushrod reciprocation. The steering is enabled by force estimation, which triggers the servo. The conical elastomer was mounted on the tip and calibrated for shear force sensing. When the elastomer tip contacts the inner wall of the pipe bends, the induced deflection generates events captured by the event camera (Fig.~\ref{fig:explore}(b)ii). We set a threshold on the shear force change to trigger the servo at 1 Hz, producing a 30° rotation of the servo motor per trigger; successive triggers accumulate to achieve larger steering angles. Since the robot and sensor rotate, the detected force direction must be transformed into the world frame. To achieve this with low computational overhead, we use a simple quantization method~\cite{5476084} to correct the force vector based on the motor's angle, mapping the result to a left or right steering command. Using this approach, the robot successfully navigated the bend.

In addition, we evaluated steering at wall intersections, as shown in Fig.~\ref{fig:explore}(c). SandWorm travels along a wall inclined at 70° to the horizontal plane and then contacts a vertical wall that intersects its direction of motion at a 70° angle. The algorithm and steering mode are identical to the last one. Experimentally, the steering manoeuvre unfolds through four observed phases: i. nominal forward motion along the initial wall; ii. elastomer tip contacting with the new wall, triggering the servo; iii. reorientation toward the new wall followed by renewed contact detection; and iv. cyclic actuation and contact checks until the elastomer tip no longer contacts the wall. Once the servo returns to its neutral position, the robot resumes forward motion. Aided by its rotational motion, SandWorm would naturally align with the new wall, thereby completing the large‐angle steering.

Finally, we evaluated extreme locomotion in a more challenging 150 mm diameter PVC elbow with a 2000 mm curvature radius, as shown in Fig.~\ref{fig:explore}(d). The material and tightened diameter introduced significantly higher friction. These conditions caused the pushrod to stall. To mitigate this, we reduced the motor speed to 30 RPM and relied on continuous low-speed rotation, which prevented shock loads. The turn was gradual, so the steering servo remained neutral. Despite the severe friction, the robot successfully traversed the bend, maintaining forward motion at 12\% of its maximum speed. This result demonstrates the mechanical robustness and compliance of the design under extreme frictional loads.

\subsection{Pipeline Dredging}
\begin{figure}[t]
    \centering
    \includegraphics[width=\linewidth]{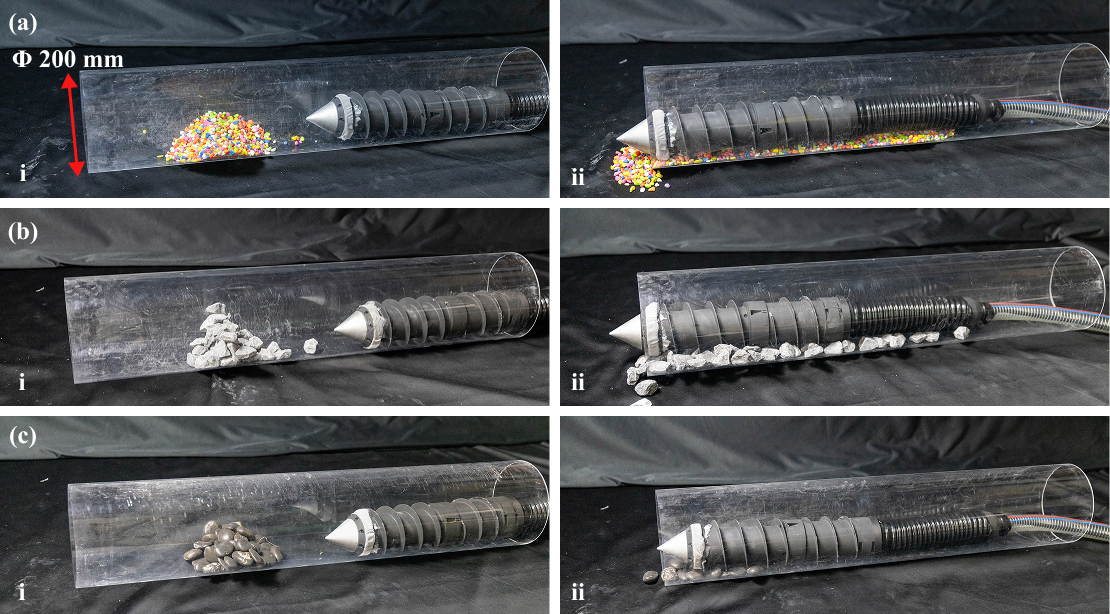}
    \caption{Pipeline dredging experiments with the autonomous SandWorm robot. (a) Gravel blockage. (b) Cobble blockage. (c) Eggstone blockage. i. image before pipe dredging, ii. image after pipe dredging. }
    \label{fig:dredge}
\end{figure}

Clearing obstructions serves as a capstone experimental validation, demonstrating the integrated capabilities of the SandWorm robot. It combines the previously characterized locomotion, perception, and mechanical design. We tested the robot's dredging performance against three blockage types: gravel, cobble, and eggstone as shown in Fig.~\ref{fig:dredge}. Each blockage spanned 200 mm and occupied approximately half the pipe's cross-section. SandWorm advanced while actively clearing the obstruction, successfully traversing the 600 mm path in 84 s, 90 s, and 81 s for each case, respectively. Compared to the 48 s unobstructed run, the speed decreased by approximately 44\%, yet the robot consistently cleared and traversed all three blockage types.

SandWorm’s high‐torque drive and threaded housing fragmented the blockage and then redistributed debris along the pipe, like a tunnel‐boring machine. The helical shell ensured uniform debris spreading and even backward transport. During dredging, SWTac captures the event stream continuously, which is processed by our stone classification network. A majority vote over sequential frames enabled robust and accurate identification of blockage material throughout the traversal. This confirms that SandWorm can not only physically dredge a pipeline but also simultaneously perceive its environment, validating the complete system integration in a challenging, application-focused scenario.

\begin{figure}[t]
    \centering
    \includegraphics[width=\linewidth]{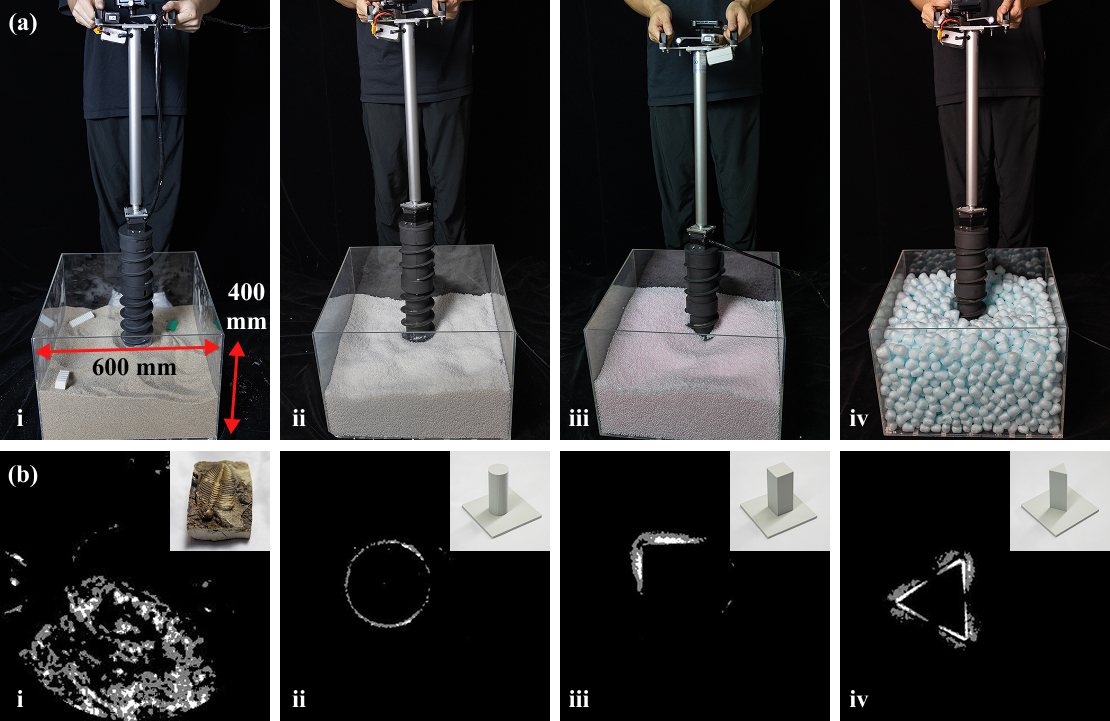}
    \caption{Subsurface exploration in granular media by SandWorm. (a) Manual experiments in: i. sand, ii. TPE, iii. EPP, iv. EPE. (b) Shapes and imaging of detected objects: i. trilobite fossil, ii. cylinder, iii. cuboid, iv. triangular prism.}
    \label{fig:detection}
\end{figure}

\begin{figure}[t]
    \centering
    \includegraphics[width=\linewidth]{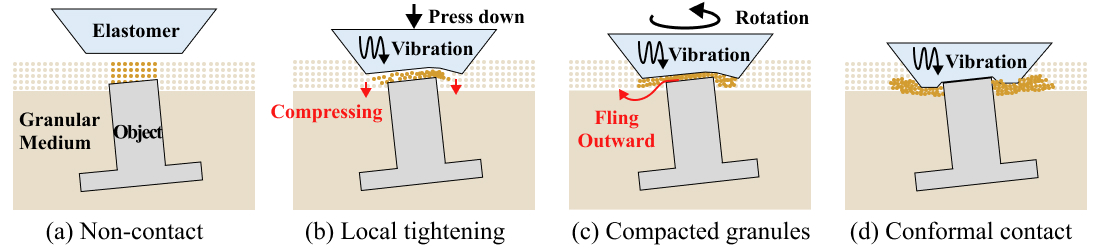}
    \caption{Sequence of elastomer-object contact in granular media, showing (a) no contact, (b) blurring from initial pressing, (c) granule compaction, and (d) granule evacuation via vibration and rotation to reveal clear object edges.}
    \label{fig:exp}
\end{figure}

\subsection{Subsurface Exploration}

For subsurface exploration, we use the manual version of SandWorm, replacing the tail module with a holding pushrod for downward force while retaining rotational drilling, as illustrated in Fig.~\ref{fig:detection}(a). The frustoconical elastomer was chosen for texture perception at the tip. The event stream was reconstructed, passed through a 3$\times$3 median filter, and visualized in real-time at 200 Hz. A smartphone interface provided communication and allows the operator to control the rotation speed and pushrod actuation.

We evaluated the object-searching performance in a 600$\times$600$\times$400 mm$^3$ test box with four media: sand, thermoplastic elastomer (TPE), expanded polypropylene (EPP), and expanded polyethylene (EPE), with respective densities of approximately 2000, 700, 100, and 10 kg/m$^3$. The key to this exploration is distinguishing the target object from the surrounding granular media by volume, density, and texture. As the tip penetrates, the active vibration mechanism gradually clears away granules from the object's surface. This allows the tip to make direct contact, revealing the object's true shape and texture rather than the granular pattern. The process is illustrated in Fig.~\ref{fig:exp}.

\begin{figure*}[t]
    \centering
    \includegraphics[width=\linewidth]{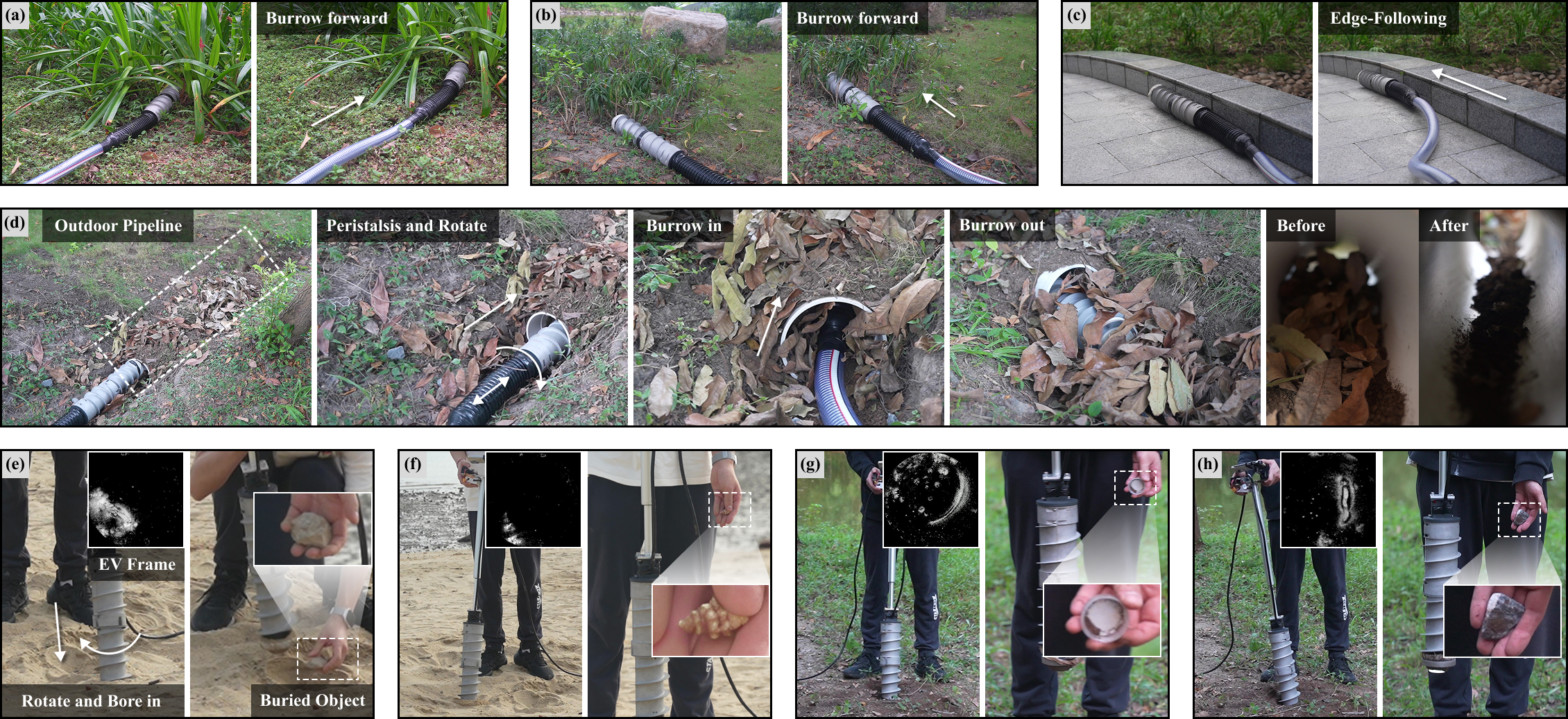}
    \caption{Field experiments demonstrating SandWorm's capabilities in real-world environments. (a) Moving forward in grass. (b) Moving forward in a bush. (c) Moving forward on a cement road. (d) Dredging in a blocked outdoor pipeline. (e)(f) Subsurface exploration experiment on the beach. (g)(h) Subsurface exploration experiment in composite garden soil. Full experimental footage is available in the supplementary video.}
    \label{fig:outdoor}
\end{figure*}

In 40 trials, operators achieved a 90\% success rate (36/40) in locating buried objects within 120 s. Sample imaging results are shown in Fig.~\ref{fig:detection}(b). As predicted by the mechanism, searching in dense sand and TPE was slower; all four trials exceeding 120 s occurred in these media. These failures in dense media occurred because high inertial torque caused the robot body to pivot, shifting the contact point to the elastomer's conical side. This resulted in reduced localization accuracy, which can be mitigated by reducing rotation speed, though at the cost of slower penetration.

\subsection{Field Validation}

To validate SandWorm's real-world performance, we conducted a final series of experiments in unstructured outdoor environments, as shown in Fig.~\ref{fig:outdoor}. These tests were designed to evaluate the system's integrated performance in complex, unpredictable conditions, where engineering improvements such as SandWorm's metal shell and the elastomer's dust-proof mesh proved crucial for durability.

First, we assessed SandWorm's locomotive performance on challenging natural terrains. SandWorm successfully traversed dense grass (Fig.~\ref{fig:outdoor}(a)), tangled bushes (Fig.~\ref{fig:outdoor}(b)), and hard cement surfaces (Fig.~\ref{fig:outdoor}(c)), demonstrating the robustness of its rotational-peristaltic locomotion. We then validated its applications in pipeline dredging. The robot was deployed in a blocked outdoor pipe, shown in Fig.~\ref{fig:outdoor}(d). Upon entering, SandWorm continuously burrowed in, actively clearing internal blockages composed of mud, leaves, and gravel. Throughout this process, SWTac indicated no prohibitive shear forces, allowing the robot to autonomously continue the clearing operation and traverse the pipe.

For subsurface exploration, we conducted numerous trials in complex, non-uniform media. The operator uses SandWorm to bore in at different locations until the target is detected in the reconstructed tactile images, as shown in Fig.~\ref{fig:outdoor}(e). We show four representative successes in Fig.~\ref{fig:outdoor}(e-h). SandWorm successfully located diverse buried objects, including a stone and a conch on a beach, as well as a bottle cap and another stone in composite garden soil. These field tests confirm that SandWorm's integrated locomotion and active perception are effective and generalizable for its target applications.

\section{Discussion}
The SandWorm robot addresses key challenges in granular media environments that traditional perception and locomotion methods face, such as low resolution, poor robustness, and vulnerability to strong vibration. By integrating a bio-inspired rotational-peristaltic locomotion mechanism with a novel event-based visuotactile sensor, SandWorm achieves efficient movement and high-resolution perception, demonstrating clear advantages in stability and adaptability over conventional approaches. Ultimately, this work demonstrates that vibration can be engineered into a perception advantage when hardware and algorithms are co-designed from first principles.

\textbf{Scientific Contributions.} This work makes three core contributions: (1) We demonstrate that active vibration enables event cameras to effectively capture static contact. By vibrating the elastomer tip, SWTac converts static contact into detectable dynamic events, enabling pixel-level tactile imaging at 1000~Hz without motion blur. This approach fundamentally expands event-based vision beyond dynamic scenes.
(2) We introduce {hardware-level perception-actuation integration}, where the same vibration drives both imaging and material penetration, while a dedicated isolation mechanism decouples and protects the sensor.
(3) We propose a bio-inspired locomotive structure combining rotation and peristalsis, which mimics marine polychaetes to achieve up to 62\% faster traversal than screw actuation alone.

\textbf{Algorithmic Innovations.} This work provides innovations in two key areas: (1) While static tactile imaging by event camera is enabled by our hardware design, we further optimized the temporal continuity of event-based perception through algorithmic solutions. We proposed IMU-guided temporal filter that exploits the correlation between elastomer acceleration and imaging quality, improving MSNR by 24\% in average with 1~ms latency.
(2) We developed algorithms for key downstream applications, including recovering full contact surfaces by leveraging asymmetric edge deformation with FEA, force estimation, and material classification.

\textbf{Engineering Insights.} This work illuminates several lessons for extreme-environment robotics: (1) Event cameras excel when vibration is harnessed as signal, not noise. SWTac transforms vibration from an interference source into an advantage, extending the application of event cameras to static tactile perception.
(2) Multimodal fusion of event-based vision with IMU are essential for maintaining signal fidelity from high-sample-rate sensors under high mechanical loads, applicable to any contact-rich task.

\textbf{Limitations and Future Work.} Despite these advances, particle adhesion to the elastomer surface may impair texture detection in sticky soils. Future work will explore tuning the elastomer's surface properties via hydrophobic coatings or micro-texturing, as well as active cleaning mechanisms (e.g., air jets, ultrasonic agitation). Additionally, extending contact surface estimation to 3D reconstruction would enable richer geometric reasoning, which can be adopted for real-time adaptive control to improve autonomous performance significantly. These steps represent a clear path toward reliable robotic perception for advanced applications in disaster rescue, archaeological excavation, and planetary exploration.

\section{Conclusion}

In this paper, we introduce SandWorm, a peristalsis-augmented screw-actuated snakelike robot, with SWTac, an event-based visuotactile sensor integrated with active elastomer vibration. By mechanically isolating the event camera, SWTac enables consistent tactile imaging in pixel-level at 1000 Hz, overcoming event cameras' intrinsic limitation of capturing only dynamic changes. For the algorithm, we proposed the IMU-guided temporal filter to enhance the consistency of the asynchronous event stream by approximately 24\%. Moreover, we verified the asymmetric edge feature by FEA and adopted the U-Net model for contact surface estimation, as well as systematically characterized sensor performance. Experimentally, SWTac provided precise force estimation (0.15 N), texture recognition (0.2 mm resolution), and stone classification (98\% accuracy). The SandWorm robot demonstrated effective autonomous locomotion, achieving straight runs (12.5 mm/s), turns, high-curvature navigation. Field trials in real-world scenarios, including pipeline dredging, subsurface exploration in beach and composite soil, further demonstrate the system's practical effectiveness.

%\clearpage

\appendices

\section{Derivation of Locomotion}
\label{app:kinematic-derivations}

This appendix provides the mathematical derivations for the locomotion components referenced in Section~\ref{sec:kinematics}. First, the screw mechanism converts the external shell's rotation $\theta$ into an axial displacement $l$ along the robot’s intrinsic axis. This relationship is given by:
\begin{equation}
l = \frac{p}{2\pi}\theta,
\label{eq:app-l}
\end{equation}
where $p$ is the pitch of the screw.

Second, we derive the effective gravitationally resolved driving component, $F_{\text{propel}}$. We consider the robot on an inclined plane with inclination $\alpha$, mass $m$, and under gravitational acceleration $g$. An axial displacement $\text{d}l$ induces a vertical height change $\text{d}h = \text{d}l \sin\alpha$. The corresponding change in gravitational potential energy $\text{d}U$ is:
\begin{equation}
\text{d}U = mg\text{d}h = mg\sin\alpha \text{d}l.
\label{eq:app-dU}
\end{equation}

$F_{\text{propel}}$ is defined as the potential energy gradient along the robot's intrinsic axis. Taking the derivative of $U$ with respect to $l$ from Eqn.~(\ref{eq:app-dU}) yields this gravitational component, which is used in the net axial force analysis in Eqn.~(\ref{eq:net-extension}) and Eqn.~(\ref{eq:net-retraction}):
\begin{equation}
F_{\text{propel}} = \frac{\text{d}U}{\text{d}l} = mg\sin\alpha.
\label{eq:app-Fpropel}
\end{equation}

\begin{figure}[t]
    \centering
    \includegraphics[width=\linewidth]{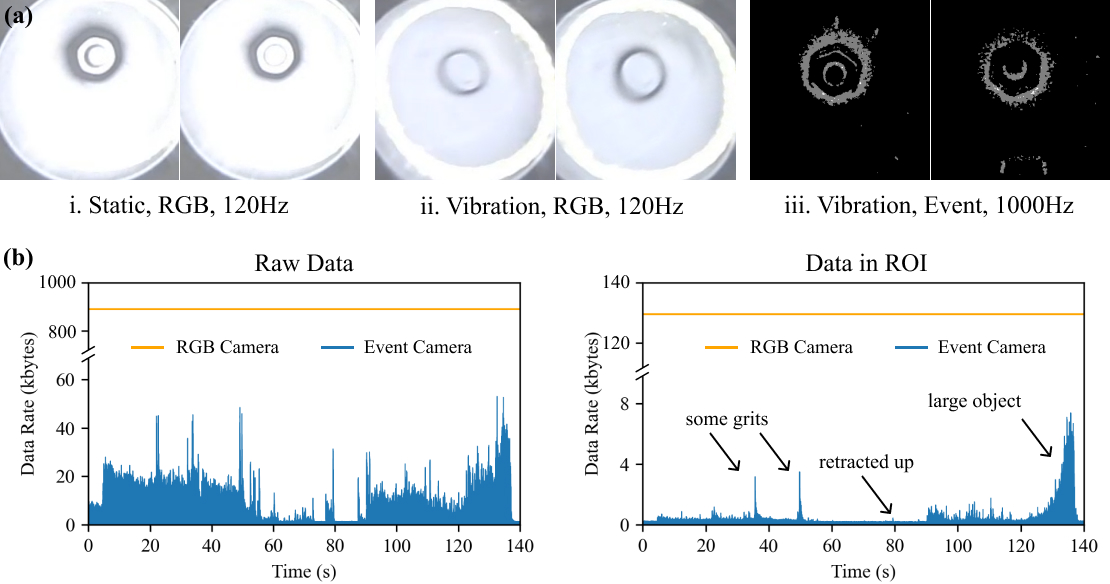}
    \caption{Event Camera vs. RGB Camera. (a) Intuitive illustration of frames. (b) Data rate comparison during the field test on the beach.}
    \label{fig:rgb}
\end{figure}

\section{Camera Comparison}

We conducted a direct comparison with an RGB camera (OV2710) at 120~Hz frame rate with exposure time fixed at $1.6$ ms, as shown in Fig.~\ref{fig:rgb}. While the RGB camera produces clear images in static contact, it suffers from significant motion blur and distortion underthe  active vibration required for our perception method. Although further reducing the exposure could mitigate the blurring, it would lead to underexposure and electronic noise, degrading the signal quality. In contrast, the event camera robustly remains sharp even when reconstructed at 1000~Hz under high-frequency vibrations. Its asynchronous pixel-level change detection enables a response latency as low as 15~$\mu s$~\cite{4444573}, serving as an effectively ultra-short exposure. This surpasses traditional photometric integration in sensitivity to rapid illumination changes, ensuring robust tactile perception with high temporal resolution.

We also performed a quantitative analysis of the data rate~\cite{funk_evetac_2024}. Using an event stream captured during a field test on the beach for comparison, we found that even at the moment of peak event generation while identifying an object, the event camera is vastly more data-efficient. This data efficiency is critical for reducing the computational and transmission loads on a computationally constrained mobile robot.

\section{Elastomer Design}
\label{app:fabrication}

\begin{figure}[t]
    \centering
    \includegraphics[width=\linewidth]{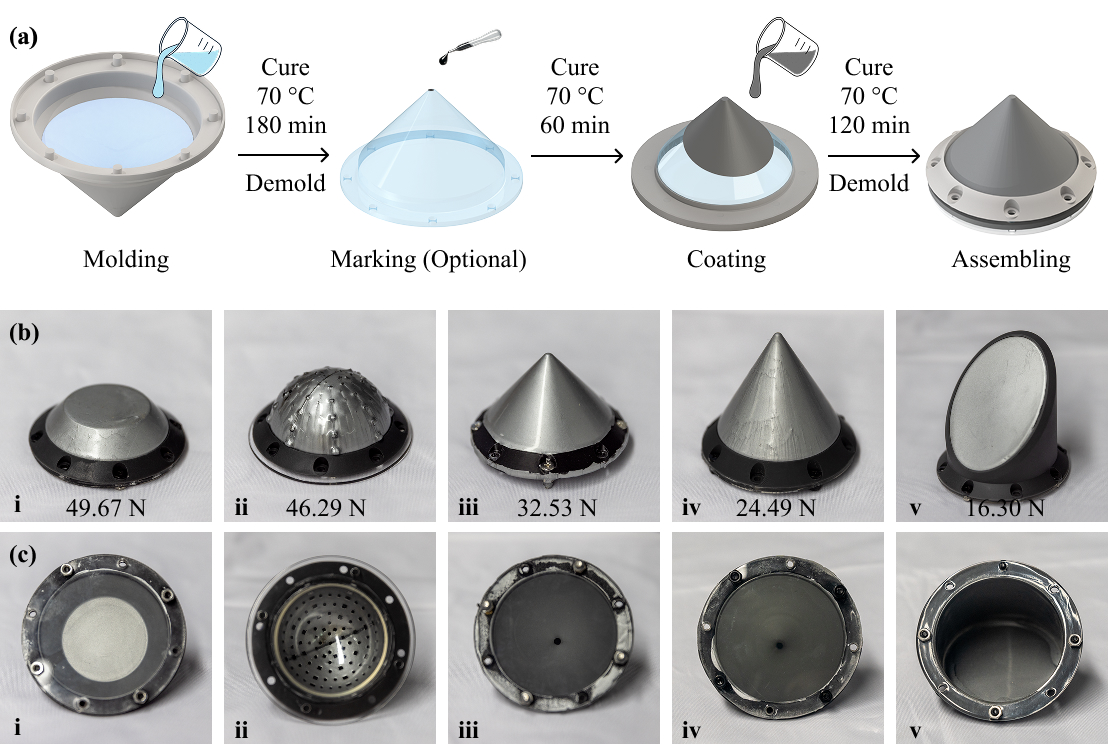}
    \caption{Fabrication of different elastomers. 
    (a) Schematic of the elastomer fabrication procedure. 
    (b) Photograph of the moulded elastomer, annotated with maximum contact force values.
    (c) Interior (imaging) side of the elastomer.}
    \label{fig:silica}
\end{figure}

The elastomer fabrication procedure is illustrated in Fig.~\ref{fig:silica}(a). PDMS base and curing agent are mixed at the target ratio, degassed under vacuum, cast into a customized mold, cured at 70 °C for 180 min, and then demolded. Next, markers are deposited by applying a thin PDMS–carbon powder composite, which is cured at 70 °C for 60 min. A reflective layer of PDMS doped with silver powder is then spin-coated over the elastomer, using a temporary insert to fill any gaps, and subsequently cured at 70 °C for 120 min. Finally, the elastomer is assembled with an acrylic substrate and mechanical latch ({P1 and P3} in Fig.~\ref{fig:structure}(e)) to protect the transparent bottom surface from contamination. To achieve different hardness levels, PDMS samples were cast with 10:1 and 17:1 base-to-curing-agent mass ratios, yielding approximate Shore hardness values of 40 A and 20 A, respectively. A 10 A hardness is realized using silicone (Ecoflex™ 00-10, Smooth-On Inc.). To evaluate long-term durability, we conducted a 15,000-cycle rotational wear test. Based on the results, we enhanced the durability by adding a thin, abrasion-resistant TPU film to the elastomer surface.
% tpu耐久，测试结果

% \subsection{Durability}
% 1
% \begin{figure}[t]
%     \centering
%     \includegraphics[width=\linewidth]{figs/wear.jpg}
%     \caption{F}
%     \label{fig:wear}
% \end{figure}

% \subsection{Geometry}

Different shapes of elastomer can be used for different functions~\cite{Shah2021VisionBasedTactileSensors}. To enhance sensor performance under different contact conditions, five elastomer geometries were fabricated. The external surfaces and the internal marker distributions are shown in Fig.~\ref{fig:silica}(b) and (c), respectively. These geometries include hemispherical designs, which incorporate a dense array of markers to achieve a wide dynamic range of force perception, and obliquely cut cylindrical geometries, which combine extended lateral surfaces with a single inclined marker line designed to support efficient drilling in granular media. Conical tips feature a single central marker and tunable aspect ratios to provide graded force sensitivity. Finally, frustoconical designs provide a flat terminal surface and broad contact area, enhancing texture discrimination. Furthermore, we experimentally validated these elastomer geometries by testing the maximum normal force experienced during penetration into and retraction from the same granular medium under identical conditions. The resulting values are annotated below each geometry in Fig.~\ref{fig:silica}(b).

\bibliographystyle{IEEEtran}
\bibliography{ref}
\vspace{-10mm}
\begin{IEEEbiography}[{\includegraphics[width=1in,height=1.25in,clip,keepaspectratio]{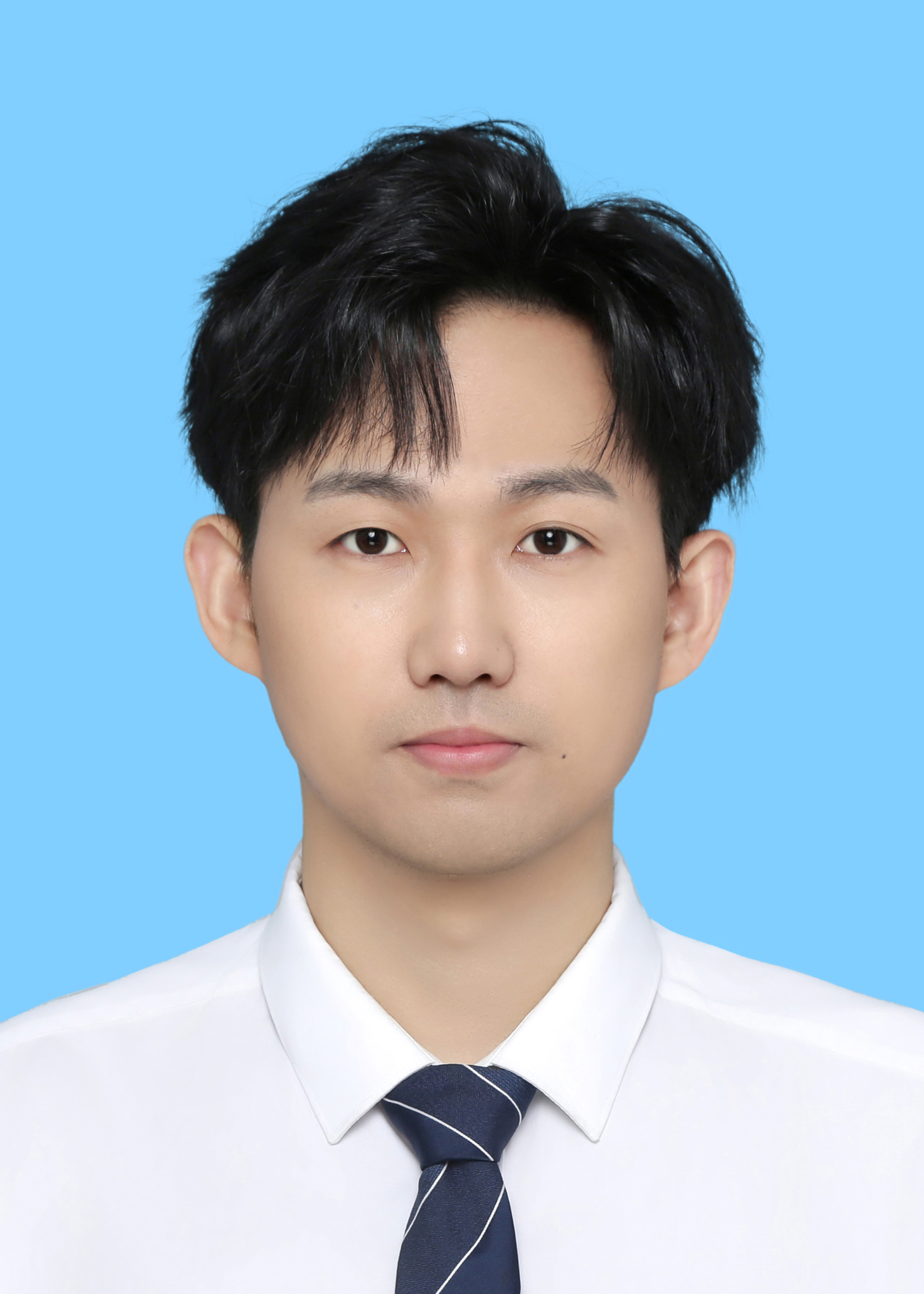}}]{Shoujie Li}
received the B.Eng. degree in electronic information engineering from the College of Oceanography and Space Informatics, China University of Petroleum,
Tsingtao, China, in 2020. He received Ph.D. degree from Tsinghua Shenzhen International Graduate School, Tsinghua University, Shenzhen, China, in 2025. He is working as a research fellow at Nanyang Technological University.

His research interests include tactile perception, grasping, and machine learning. He received the Outstanding Mechanisms and Design Paper Finalists in ICRA 2022 and the Best Application Paper Finalists in IROS 2023. He won first place in the Robotic Grasping of Manipulation Competition-Picking in Clutter in ICRA 2024. 
\end{IEEEbiography}

\vspace{-10mm}

\begin{IEEEbiography}[{\includegraphics[width=1in,height=1.25in,clip,keepaspectratio]{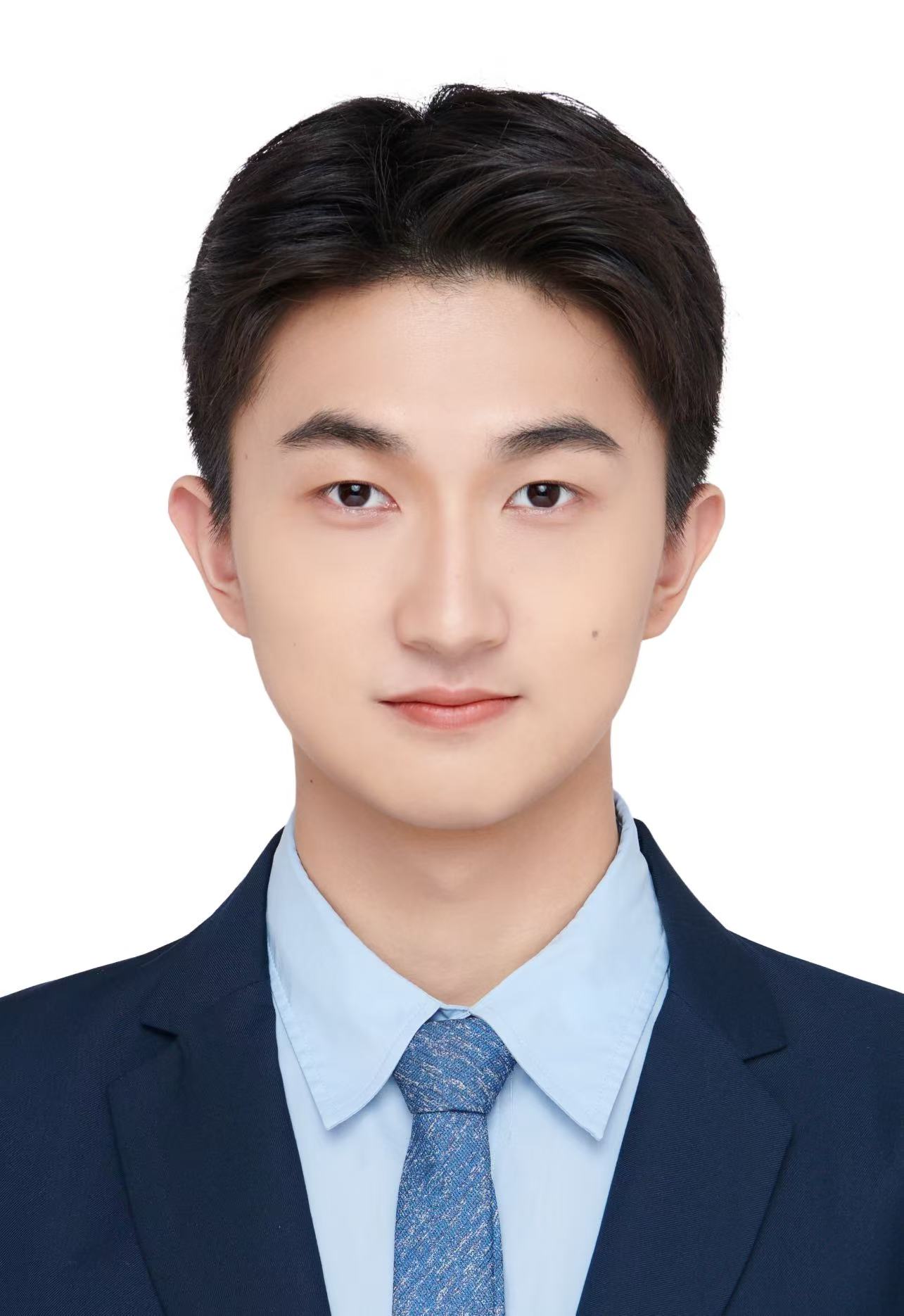}}]{Chanqing Guo} received the B.S. degree in Robotic Engineering from Shein-Ming Wu School of Intelligent Engineering, South China University of Technology, Guangzhou, China. He is currently pursuing the M.S. degree in Data Science and Information Technology at Tsinghua Shenzhen International Graduate School, Tsinghua University, Shenzhen, China.

His research interests focus on tactile perception, with particular applications in novel structure design and robot learning.

\end{IEEEbiography}

\vspace{-10mm}

\begin{IEEEbiography}[{\includegraphics[width=1in,height=1.25in,clip,keepaspectratio]{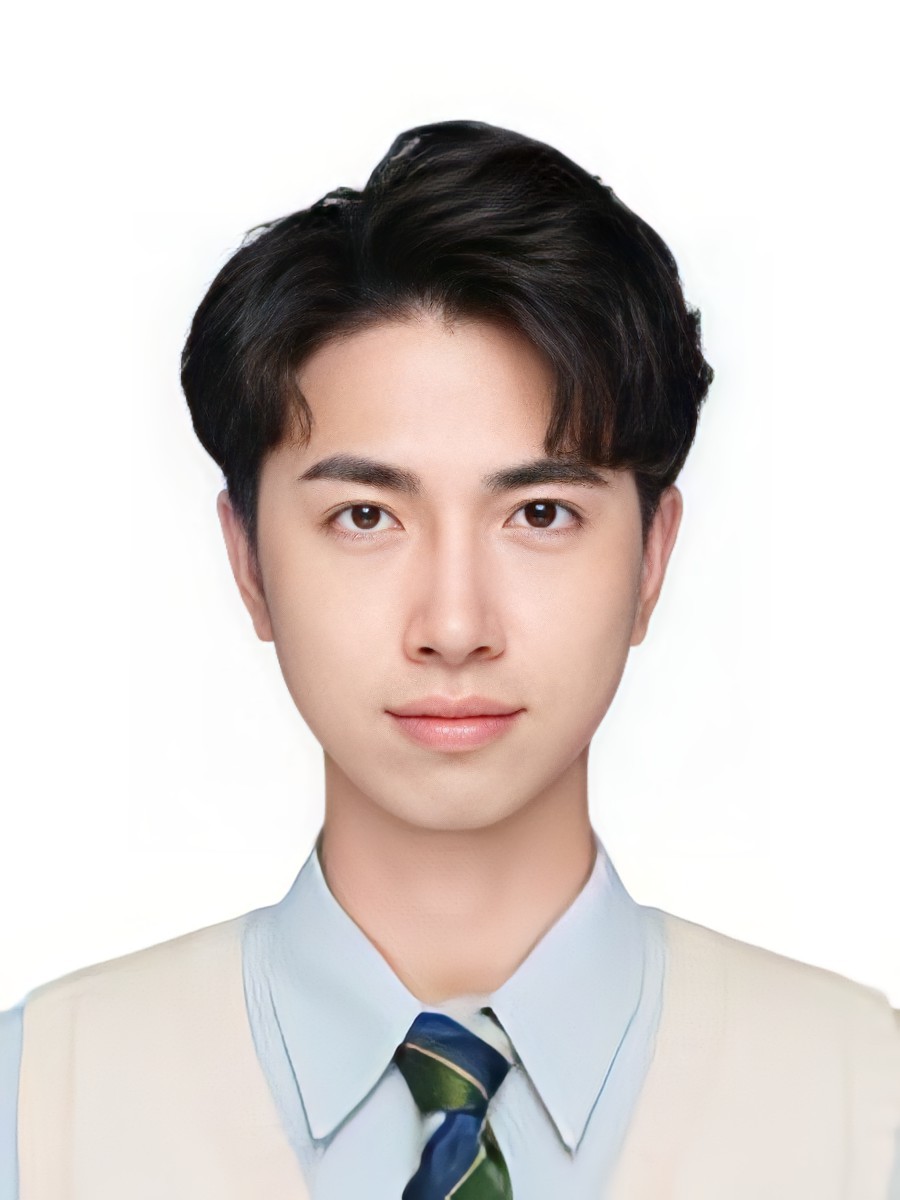}}]{Junhao Gong} received the B.S. degree in Electronic Information Engineering from Harbin Institute of Technology, Weihai, China, in 2024. He is currently pursuing his Ph.D. degree in Tsinghua Shenzhen International Graduate School, Tsinghua University, Shenzhen, China.
His research interests mainly focus on tactile sensing and contact-aware robotic manipulation.

\end{IEEEbiography}

\vspace{-10mm}

\begin{IEEEbiography}[{\includegraphics[width=1in,height=1.25in,clip,keepaspectratio]{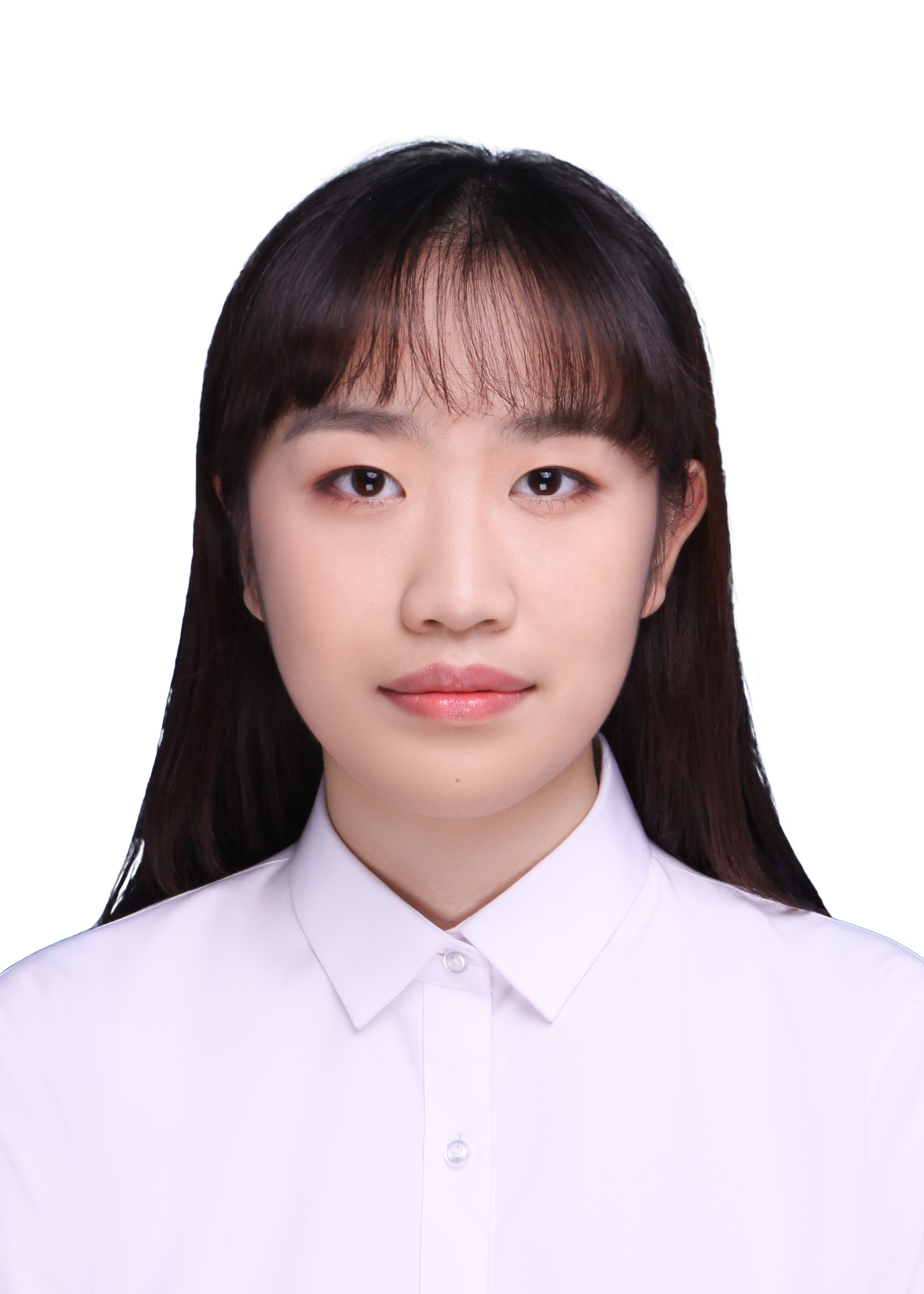}}]{Chenxin Liang}  received the B.S. degree in Electronic Engineering from Tianjin University, Tianjin, China, in 2023. She is currently pursuing the Ph.D. degree with the Tsinghua Shenzhen International Graduate School, Tsinghua University, Shenzhen, China. 

Her research interests include vision-based tactile sensing, 3D reconstruction, and depth estimation.

\end{IEEEbiography}

\vspace{-10mm}

\begin{IEEEbiography}[{\includegraphics[width=1in,height=1.25in,clip,keepaspectratio]{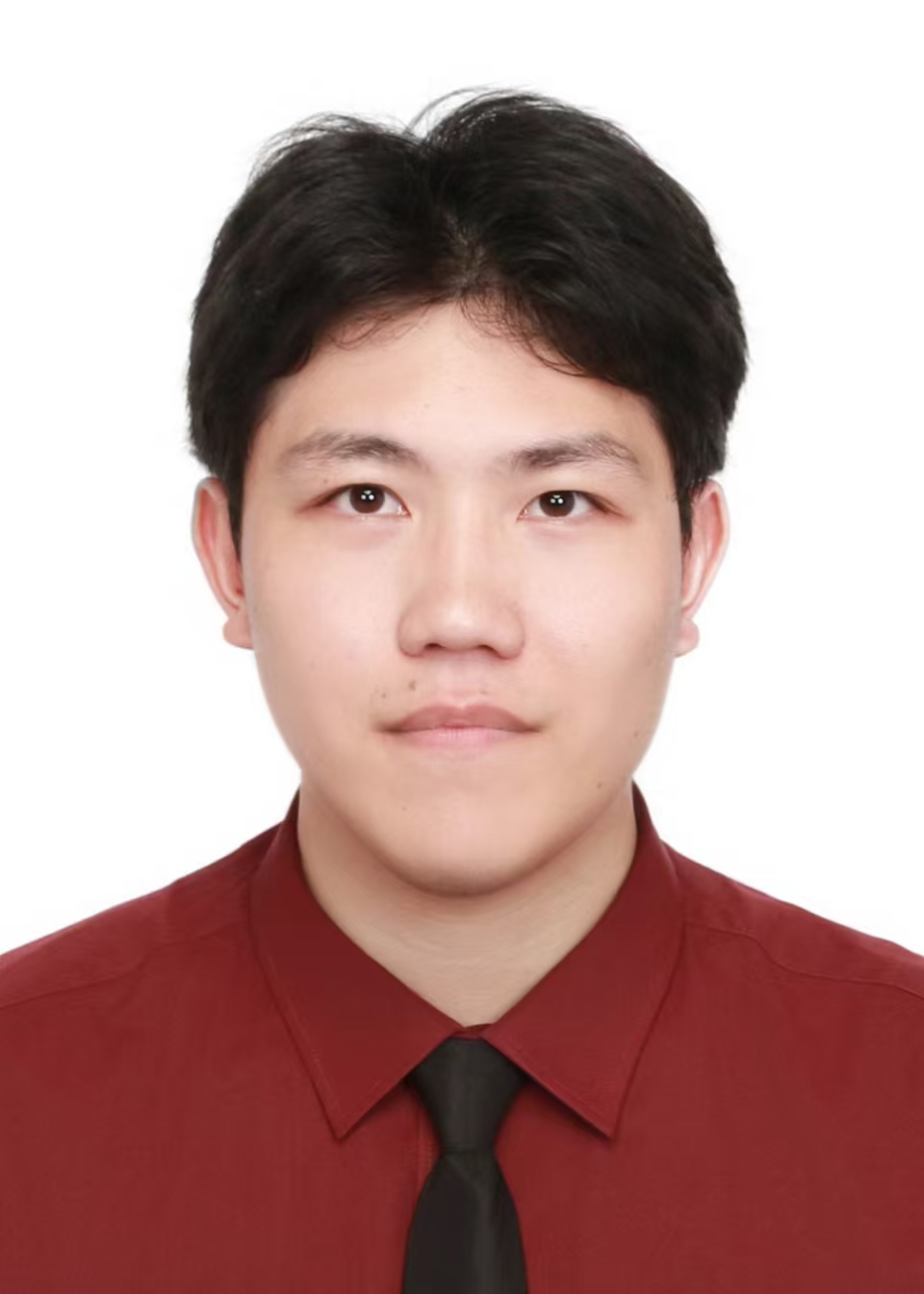}}]{Wenhua Ding}  received the B.E. degree from the Shine-Ming Wu School of Intelligent Engineering, South China University of Technology, China, in 2024. He is currently pursuing the Master’s degree at the Tsinghua Shenzhen International Graduate School, Tsinghua University, China. 

His research interests include robotics and UAV-related sensing technologies.
\end{IEEEbiography}

\vspace{-10mm}

\begin{IEEEbiography}
[{\includegraphics[width=1in,height=1.25in,clip,keepaspectratio]{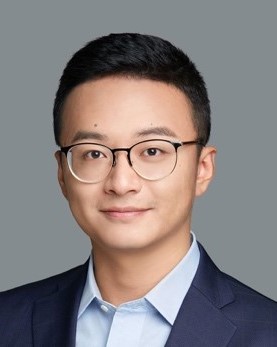}}]{Wenbo Ding}
received the BS and PhD degrees (Hons.) from Tsinghua University in 2011 and 2016, respectively. He worked as a postdoctoral research fellow at Georgia Tech from 2016 to 2019. He is now an associate professor and PhD supervisor at Tsinghua Shenzhen International Graduate School, Tsinghua University, where he leads the Smart Sensing and Robotics (SSR) group. His research interests are diverse and interdisciplinary, which include tactile sensing, embodied intelligence and AI4S. 

He has received many prestigious awards, including the IROS New Generation Star, Gold Medal of the 47th International Exhibition of Inventions Geneva and the IEEE Scott Helt Memorial Award. He serves as the editorial board member and co-chair/area chair for several top journals/conferences.
\end{IEEEbiography}

\end{document}